%% file: main.tex
\documentclass[10pt,twocolumn,letterpaper]{article}

\usepackage{cvpr}              
\usepackage{comment}

\usepackage{color} 
\usepackage{multirow} 
\usepackage{amsmath, amsfonts, amsthm, amssymb}
\usepackage[inline]{enumitem}
\usepackage{booktabs}

\usepackage[accsupp]{axessibility}  

\usepackage[switch]{lineno} 

\usepackage{colortbl}
\usepackage{xcolor}
\usepackage{tcolorbox}

\newtheoremstyle{mysubtask}%
{}{}{\normalfont}{}{\scshape}{.\,}{ }{}
\theoremstyle{mysubtask}

\newtheoremstyle{mytask}%
{}{}{\normalfont}{}{\scshape}{.\,}{ }{}
\theoremstyle{mytask}

\newtheoremstyle{mydef}%
{}{}{\normalfont}{}{\scshape}{.\,}{ }{}
\theoremstyle{mydef}

\usepackage{xcolor}
\usepackage{soul}

\definecolor{mygreen}{rgb}{0.286,0.737,0.392}
\definecolor{bblue}{HTML}{4F81BD}
\definecolor{oorange}{HTML}{F4C842}
\definecolor{rred}{HTML}{C0504D}
\definecolor{ggreen}{HTML}{9BBB59}
\definecolor{ppurple}{HTML}{9F4C7C}
\definecolor{darkgreen}{HTML}{228B22}
\definecolor{cred}{HTML}{D81B60}
\definecolor{cblue}{HTML}{1E88E5}
\definecolor{cyellow}{HTML}{FFC107}
\definecolor{nred}{HTML}{e41a1c}
\definecolor{nblue}{HTML}{377eb8}
\definecolor{ngreen}{HTML}{4daf4a}
\definecolor{lblue}{HTML}{6C8EBF}
\definecolor{lightgray}{rgb}{0.95, 0.95, 0.95}
\definecolor{amber}{rgb}{1.0, 0.75, 0.0}

\newcommand{\hgrow}{\rowcolor{lightgray!70}}

\usepackage{pifont}
\newcommand{\xmark}{\textcolor{red}{\ding{55}}}
\newcommand{\cmark}{\textcolor{darkgreen}{\ding{51}}}

\usepackage{arydshln}
\usepackage{siunitx}


\definecolor{cvprblue}{rgb}{0.21,0.49,0.74}
\usepackage[pagebackref,breaklinks,colorlinks,allcolors=cvprblue]{hyperref}

\def\paperID{14973} 
\def\confName{CVPR}
\def\confYear{2025}

\title{\includegraphics[width=0.7cm, height=0.7cm]{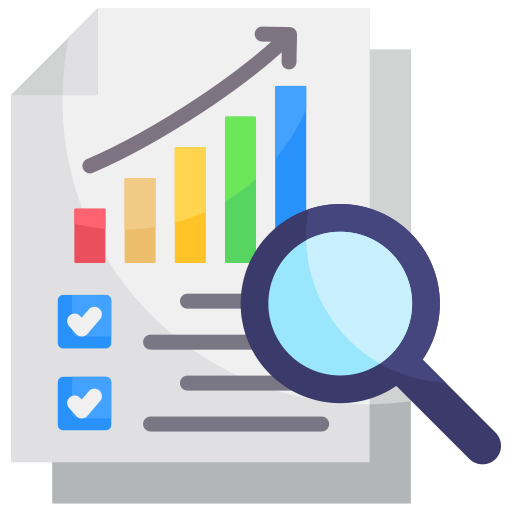}VDocRAG: Retrieval-Augmented Generation over Visually-Rich Documents}

\author{
    Ryota Tanaka\textsuperscript{1,2} \quad Taichi Iki\textsuperscript{1} \quad Taku Hasegawa\textsuperscript{1} \quad Kyosuke Nishida\textsuperscript{1} \quad Kuniko Saito\textsuperscript{1} \quad Jun Suzuki\textsuperscript{2} \\
$^{1}$NTT Human Informatics Laboratories, NTT Corporation \quad ${^2}$Tohoku University \\
    \href{https://vdocrag.github.io}{https://vdocrag.github.io}
}

\begin{document}
\maketitle

\begin{abstract}
We aim to develop a retrieval-augmented generation (RAG) framework that answers questions over a corpus of visually-rich documents presented in mixed modalities (e.g., charts, tables) and diverse formats (e.g., PDF, PPTX). In this paper, we introduce a new RAG framework, VDocRAG, which can directly understand varied documents and modalities in a unified image format to prevent missing information that occurs by parsing documents to obtain text. To improve the performance, we propose novel self-supervised pre-training tasks that adapt large vision-language models for retrieval by compressing visual information into dense token representations while aligning them with textual content in documents. Furthermore, we introduce OpenDocVQA, the first unified collection of open-domain document visual question answering datasets, encompassing diverse document types and formats. OpenDocVQA provides a comprehensive resource for training and evaluating retrieval and question answering models on visually-rich documents in an open-domain setting. Experiments show that VDocRAG substantially outperforms conventional text-based RAG and has strong generalization capability, highlighting the potential of an effective RAG paradigm for real-world documents. \end{abstract}

\section{Introduction}
Large language models (LLMs) have demonstrated impressive performance on diverse natural language tasks~\cite{suzgun2022challenging,achiam2023gpt,jiang2024mixtral,dubey2024llama}. These models struggle with factual errors despite their increased model and data scale~\cite{mallen-etal-2023-trust,maekawa-etal-2024-retrieval}. To remedy this problem, retrieval-augmented generation (RAG) methods~\cite{lewis2020retrieval,guu2020retrieval} can retrieve knowledge from an external corpus, potentially reducing hallucination and increasing knowledge coverage.  Most previous RAG frameworks assume the context is composed entirely of text, with no graphical elements. In contrast, a significant amount of real-world information is stored in visually-rich documents, such as charts, tables, web pages, and office documents. These documents often contain both textual and visual objects, with content spread structurally across various locations depending on diverse formats and types.

\begin{figure}
    \centering
    \includegraphics[width=.47\textwidth]{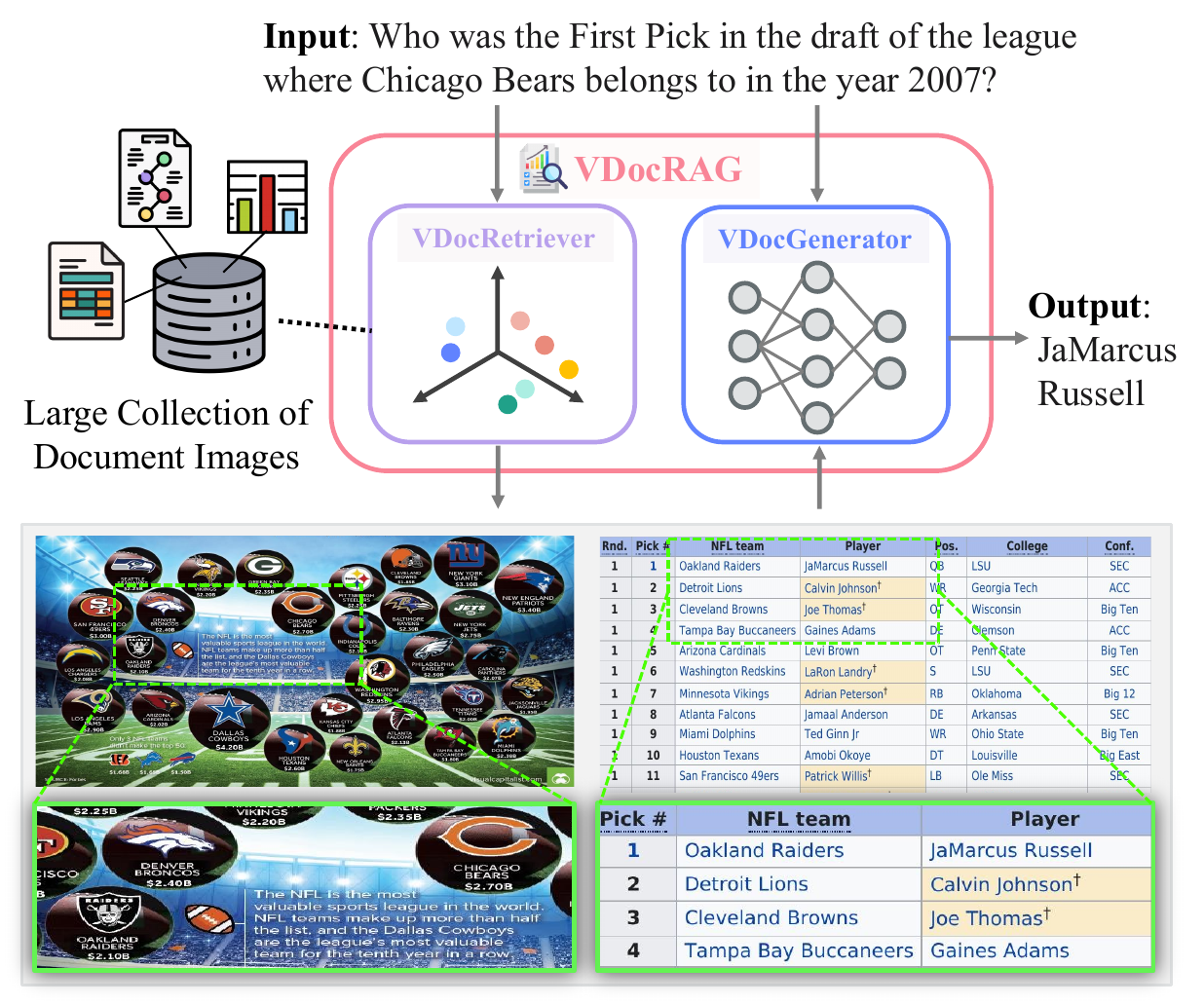}
    \caption{Our framework of VDocRAG and examples from OpenDocVQA. VDocRAG consists of VDocRetirver and VDocGenerator, which can retrieve relevant documents and generate answers by understanding the original appearance of documents.}
    \label{fig:intro}
\end{figure}

Thus, document visual question answering (DocumentVQA)~\cite{Mathew_2021_WACV,DBLP:conf/aaai/TanakaNY21,Mathew_2022_WACV,SlideVQA2023} aims to build an agent capable of reading and comprehending document images to answer the question. Here, most existing DocumentVQA questions operate in a closed setting without requiring any retrieval. While this definition simplifies the QA model, it does not reflect many real-world use cases where the question is asked through some open-domain natural language interface, such as QA systems searching information across in-house documents or customer service chatbots on e-commerce websites. To address this limitation, recent works have introduced retrieval tasks on document images~\cite{faysse2024colpali,ma2024unifying}. However, these cannot fully develop models that effectively integrate the retrieved information into the final output. This gap hinders the application of DocumentVQA models in more realistic, open-domain scenarios.

In this paper, we introduce a new RAG framework, VDocRAG, which can directly understand varied documents and modalities in a unified image format to avoid tedious parsing and potential information loss that occurs in conventional text-based RAG. As depicted in Figure~\ref{fig:intro}, VDocRAG consists of two main components, both of which effectively leverage the visual features of documents. First, VDocRetriever retrieves document images related to the question from a corpus of document images. Second, VDocGenerator uses these retrieved images to generate the answer. To encode document images and interact with the encoded information, we adapt pre-trained large vision language models (LVLMs)~\cite{abdin2024phi,laurenccon2024building} as the backbone for VDocRAG. Since LVLMs are inherently generative models, it is sub-optimal for embeddings as they prevent the representations from capturing information across the entire input sequence due to the training objective (i.e., next-token prediction). To bridge this gap, we introduce new self-supervised pre-training tasks that harness the understanding and generation capabilities of LVLMs to enhance representation learning. Specifically, we compress the entire image representation into a dense token representation, by aligning the text in documents via retrieval and generation tasks.

Furthermore, we introduce OpenDocVQA, the first unified collection of open-domain DocumentVQA datasets encompassing a wide range of document types and formats. OpenDocVQA provides a comprehensive resource for training and evaluating retrieval and question answering models on visually-rich documents in an open-domain setting. Experiments demonstrate that VDocRAG substantially outperforms conventional text-based RAG and has strong generalization performance. 

Our main contributions are summarized as follows:
\begin{itemize}
\item  We introduce a new RAG framework, VDocRAG, which can directly understand diverse real-world documents purely from visual features.

\item  We are the first to explore pre-training tasks designed for document retrieval-oriented adaptation of LVLMs, by compressing visual document representations.

\item We introduce OpenDocVQA, the first unified open-domain DocumentVQA dataset with diverse documents.


\end{itemize}
\section{Related Work}
\paragraph{Retrieval-augmented generation (RAG).} RAG in the NLP community aims at retrieving external knowledge to reduce factual errors and enhance performance in various knowledge-intensive tasks~\cite{borgeaud2022improving,mallen-etal-2023-trust,maekawa-etal-2024-retrieval,asai-etal-2023-retrieval,ram2023context}. Inspired by the success of RAG in NLP, this technique has also applied applications across different domains, including images~\cite{ramos2023smallcap,ramos-etal-2023-retrieval,chen2022re,yasunaga2023retrieval}, codes~\cite{parvez2021retrieval,zhou2022docprompting}, videos~\cite{chen2023retrieval,xu2024retrieval}, audio~\cite{koizumi2020audio,yang2024instructtts}, and 3D~\cite{seo2024retrieval,zhang2023remodiffuse}. However, most existing works have focused on retrieving knowledge from only plain-text documents or non-text media. In contrast, we tackle the challenge of extracting knowledge from visually-rich documents organized in complex, multimodal formats.

\paragraph{Visual document retrieval and visual RAG.} With the success of LLMs, there is a growing trend to build large vision language models (LVLMs) that integrate image understanding capabilities by combining image encoders~\cite{RaffelSRLNMZLL20,li2022blip,zhai2023sigmoid} with LLMs~\cite{liu2023llava,li2023blip2,abdin2024phi,instructblip,laurenccon2024building,tanaka2024instructdoc}. Concurrent works in visual document retrieval~\cite{ma2024unifying,faysse2024colpali,dong2025mmdocir} and visual RAG~\cite{yu2024visrag,cho2024m3docrag,ma2024visa} leverage LVLMs to directly encode visually-rich documents through images. However, these approaches have trouble understanding diverse real-world documents due to the limitations of their datasets and training strategies. The existing visual document retrieval dataset, ViDoRe~\cite{ma2024unifying}, contains questions that might not require retrieval and handles a limited number of document types, resulting in a gap between real-world scenarios. In contrast, our dataset covers open document types and provides questions that are verified by humans to require retrieval and to have context-independent conditions for the retrieval. From the perspective of training, despite the significant gap between generative pre-training tasks and retrieval tasks in LVLMs, previous works~\cite{ma2024unifying,faysse2024colpali,yu2024visrag,cho2024m3docrag,ma2024visa} leverage LVLMs without specific training for bridging the gap. To address this, we introduce pre-training tasks that transfer the understanding and generation capabilities of LVLMs to retrievers.

\paragraph{Document visual question answering (DocumentVQA).} DocumentVQA is a high-level document understanding task that involves answering questions on visually-rich documents. These documents include a variety of elements, such as handwritten and digital text~\cite{Mathew_2021_WACV,DBLP:conf/aaai/TanakaNY21}, complex layouts~\cite{landeghem2023document,zhang2023mpmqa,zhu2022towards}, and graphical elements~\cite{Mathew_2022_WACV,masry-etal-2022-chartqa,SlideVQA2023}. However, previous studies have assumed closed settings that do not require retrieval, except for Dureader$_\text{vis}$~\cite{qi-etal-2022-dureadervis}. Our work differs from Dureader$_\text{vis}$ as follows. First, OpenDocVQA covers a wide range of document formats and domains, while Dureader$_\text{vis}$ focuses on screenshots of websites, limiting its generalizability. Second, OpenDocVQA reflects more real-world scenarios that require both single- and multi-hop reasoning over documents, while Dureader$_\text{vis}$ requires only single-hop reasoning. Lastly, even lexical search methods yield sufficient performance in Dureader$_\text{vis}$ due to its reliance on textual content. In contrast, OpenDocVQA requires a visual semantic search where visual and contextual information can be exploited.

\section{OpenDocVQA Task and Dataset}
\subsection{Task Formulation}
Given a large collection of $N$ document images $\mathcal{I} = \{I_1, ..., I_N\}$ and a question $Q$, the goal of OpenDocVQA task is to output an answer $A$ by finding the relevant $k$ images $\mathcal{\hat{I}} \in \mathcal{I}$, where $k \ll N$. We decompose the task into two stages. \textbf{Visual document retrieval}: given $Q$ and $\mathcal{I}$, the model retrieves the relevant $k$ images $\mathcal{\hat{I}}$ from which to derive the answer. \textbf{DocumentVQA}: the model takes $Q$ and the retrieved images $\mathcal{\hat{I}}$ as input, to generate $A$. 

OpenDocVQA covers multiple open-domain DocumentVQA datasets with diverse document types. To reflect real-world scenarios, we evaluate models with both \textbf{single-pool} and \textbf{all-pool} settings. In the single-pool setting, retrieval is performed from a specific pool of documents provided by each original dataset. The all-pool setting requires retrieving from the entire candidate pool, which includes documents from a wide range of domains.

\begin{figure}
    \centering
    \includegraphics[width=.5\textwidth]{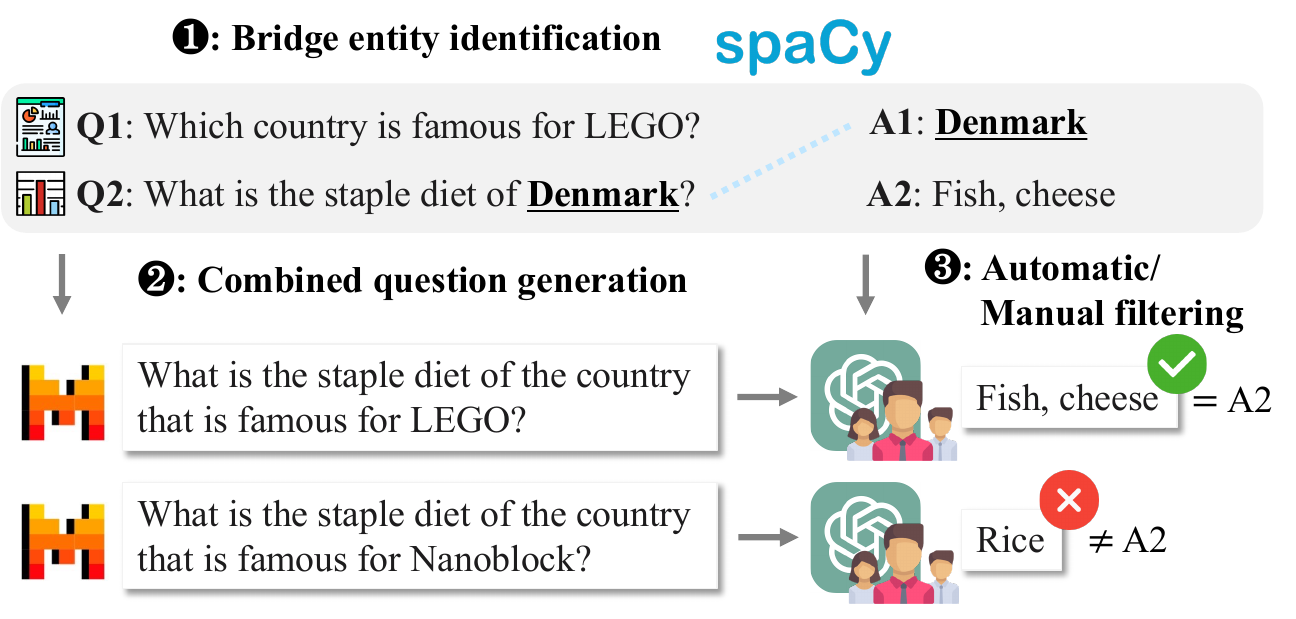}
    \caption{Process of creating multi-hop DocumentVQA questions.}
    \label{fig:multihop}
\end{figure}

\subsection{Dataset Collection}

\paragraph{Filtering of DocumentVQA datasets.}
We collected and filtered instances of seven existing document VQA datasets~\cite{Mathew_2021_WACV,DBLP:conf/aaai/TanakaNY21,masry-etal-2022-chartqa,Mathew_2022_WACV,SlideVQA2023,landeghem2023document,zhang2023mpmqa}. Most of their questions are context-\textbf{dependent} conditions, where they cannot be answered without referencing the accompanying document (e.g., \textit{What is the title?}). Therefore, we filtered out questions lacking sufficient context for retrieval. To address this, we initially applied heuristic rules to automatically select likely context-\textbf{independent} questions, reducing the pool by 20.9\%. Then, we manually reviewed and verified the remaining examples to ensure their context independence.

\paragraph{Reformulation of TableQA dataset.}
We used QA pairs from Open-WikiTable~\cite{kweon-etal-2023-open}, an open-domain TableQA dataset that required retrieving tables from Wikipedia to answer the question. Since the original dataset provides tables in only textual format (HTML data), we took the screenshot images of tables from the corresponding Wikipedia pages to reformulate the task as the OpenDocVQA.

\paragraph{Creation of new multi-hop questions.}
 To enhance the model's ability to interact with multiple document sources (e.g., charts and tables), we semi-automatically created a multi-hop DocumentVQA dataset, MHDocVQA, using the single-hop QA pairs collected in the previous steps. As shown in Figure~\ref{fig:multihop}, the creating process involved the following steps: (1) We first used spaCy~\cite{spacy2} to identify a \textit{bridge entity} (e.g., \textit{Denmark}) in the answer to a single-hop question and then searched for this entity in other single-hop questions. (2) Next, we used Mixtral-8x22B~\cite{jiang2024mixtral} to combine the two single-hop questions. (3) We filtered the generated multi-hop questions using another LLM (GPT-4o~\cite{achiam2023gpt}), which answered the questions based on the context of the two initial single-hop questions and their answers. If the predicted answer was the same as the answer to the second single-hop question, the multi-hop question was validated. Finally, we manually reviewed the filtered questions to ensure their quality before including them in our dataset. 

\paragraph{Negative candidates mining.}
We produced negative image candidates for retrievers to sift through for every question, used only during inference. We first extracted OCR text from images in the COYO-700M dataset~\cite{kakaobrain2022coyo-700m}, a web-scaled image collection. Subsequently, we mined negative images where the OCR text exhibits high lexical overlap with the question but does not contain the correct answer. 

\begin{table}[t!]
    \centering
        \scalebox{0.88}{
    \tabcolsep=2.8pt
    \small
    \begin{tabular}{lccc} 
        \toprule
         & ViDoRe~\cite{faysse2024colpali} & Dureader$_{\text{vis}}$~\cite{qi-etal-2022-dureadervis}  & OpenDocVQA \\ \midrule
        Retrieval & \cmark & \cmark & \cmark  \\
        QA & \xmark & \cmark & \cmark  \\
        Context-Independent & \xmark & \cmark & \cmark  \\
        Visual Semantic Search & \cmark  & \xmark& \cmark \\ 
        Multi-Hop & \xmark& \xmark  & \cmark  \\
        Document Contents & T, L, F, C, D & T, L & T, L, F, C, D \\
        Answer Types & -- & Ext & Ext, Abs \\
        \#Document Types & 6 & 1 & Open \\ 
        \#QAs & 3,810 & 15,000 & 43,474 \\
        \#Images (Pages) & 8,310 & 158,000 & 206,267 \\
        \bottomrule
    \end{tabular}
    }
    \caption{Comparison of related datasets. Document contents include (T)able, (L)ist, (F)igure, (C)hart, and (D)iagram. Answer
types are Extractive (Ext) and Abstractive (Abs).}
    \label{tab:comparison}
\end{table}

\begin{figure*}[t!]
    \centering
\includegraphics[width=0.95\textwidth]{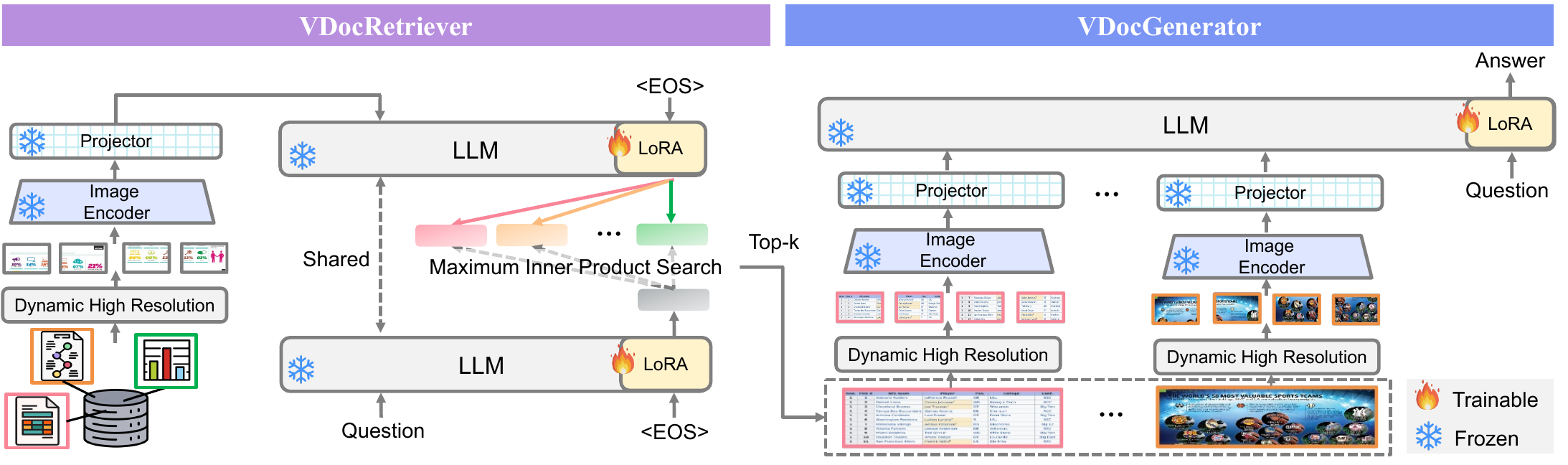}
    \caption{Overview of our VDocRAG model. VDocRetriever retrieves document images related to the question from a corpus of document images, and VDocGenerator uses these retrieved images to generate the answer.}
    \label{fig:vdocrag}
\end{figure*}

\subsection{Comparison with Related Datasets}
Table~\ref{tab:comparison} shows the statistics of OpenDocVQA and other related datasets, including ViDoRe~\cite{faysse2024colpali} and Dureader$_{\text{vis}}$~\cite{qi-etal-2022-dureadervis}. OpenDocVQA has three unique key properties: First, it is the first large-scale collection of open-domain DocumentVQA datasets to address open document types, whereas ViDoRe considers six document types for only the retrieval task and Dureader$_{\text{vis}}$ is limited to webpages. Second, the questions in OpenDocVQA are context-independent and require visual semantic search, whereas ViDoRe's questions are context-dependent, and even lexical search methods yield sufficient performance in Dureader$_{\text{vis}}$. This indicates our dataset better reflects real-world scenarios. Lastly, unlike ViDoRe and Dureader$_{\text{vis}}$, OpenDocVQA requires multi-hop reasoning with extractive (e.g., \textit{span, list}) and abstractive (e.g., \textit{arithmetic, counting, no answer}) answer types, providing a more challenging setting. 

\section{Proposed Model}

\subsection{Architecture Overview}
As shown in Figure~\ref{fig:vdocrag}, VDocRAG consists of two components: VDocRetriever and VDocGenerator. Our approach adopts the pre-trained LVLMs to unify the varied formats and modalities in a single form as an image for direct document understanding.

\paragraph{Dynamic high-resolution image encoding.}
To encode high-resolution images with various aspect ratios, a dynamic cropping~\cite{ye-etal-2023-ureader,dong2024internlm} is utilized to split the image into smaller patches while maintaining the integrity of the original aspect ratio. Each patch is a small image with $336\times336$ size, and we treat them as individual inputs for the image encoder. After encoding images, we convert them via a projector (two-layer MLP) into visual document features $\mathbf{z}_\text{d}$.

\paragraph{VDocRetriever.}
VDocRetriever is an LVLM-based dual-encoder architecture that encodes queries and document images independently. We append an \texttt{<EOS>} token to the end of the question and visual document features $\mathbf{z}_{\text{d}}$, and then feed them into the LLM to obtain the question and visual document embeddings ($\mathbf{h}_{\text{q}}$, $\mathbf{h}_{\text{d}}$) by taking the last layer \texttt{<EOS>} vector. Then, it retrieves $k$ documents $\mathcal{\hat{I}}$ with the $k$ highest similarity scores to the question. Formally, the similarity scores between the question and visual document embeddings are computed via maximum inner product search~\cite{douze2024faiss}, as follows: $\textsc{sim}(\mathbf{h}_{\text{q}}, \mathbf{h}_{\text{d}}) = \frac{\mathbf{h}_{\text{q}}^\top \mathbf{h}_{\text{d}}}{\|\mathbf{h}_{\text{q}}\| \|\mathbf{h}_{\text{d}}\|}$. 

\paragraph{VDocGenerator.} VDocGenerator adapts LVLM to generate answers $A$ given the question $Q$ and the retrieved $k$ documents $\mathcal{\hat{I}}$ obtained from VDocRetriever. After encoding the retrieval result, we concatenate the question and the encoded result, then feed this combined input into the LLM.

\begin{figure*}[t!]
    \centering
\includegraphics[width=1.0\textwidth]{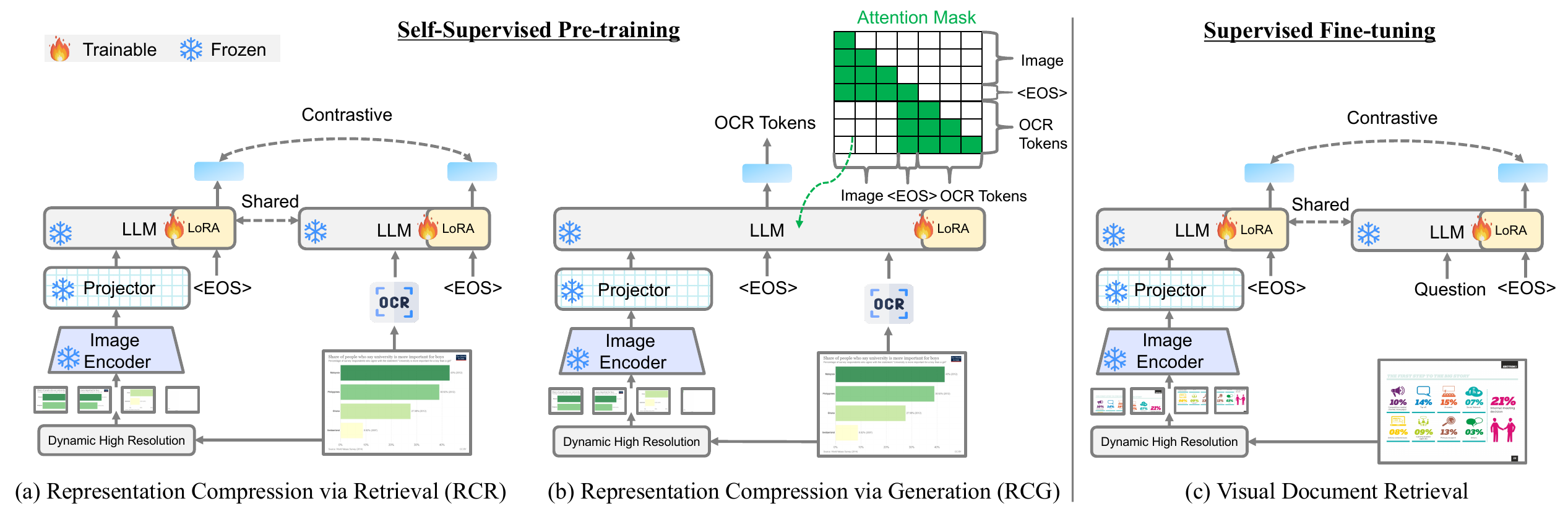}
    \caption{Our pre-training tasks using unlabeled documents and fine-tuning in VDocRetriever. The RCR task retrieves relevant images given corresponding OCR tokens, and the RCG task outputs OCR tokens by paying attention to only the \texttt{<EOS>} token.}
    \label{fig:pretrain}
\end{figure*}

\subsection{Self-Supervised Pre-training Tasks} 
Figure~\ref{fig:pretrain}a and \ref{fig:pretrain}b show our pre-taining tasks in VDocRetriever. The goal of pre-training is to transfer the powerful understanding and generation abilities of LVLMs to facilitate their usage in visual document retrieval. To this end, we propose two new self-supervised pre-training tasks to compress the entire image representation into the \texttt{<EOS>} token at the end of the input image. Our pre-training process passes the document image, and its extracted OCR text is used as a pseudo target. Full pre-training objectives is defined as $\mathcal{L} = \mathcal{L}_{\text{RCR}} + \mathcal{L}_{\text{RCG}}$.

\paragraph{Representation Compression via Retrieval (RCR).} 
We compress image representations with a contrastive learning task that retrieves images relevant to their corresponding OCR text, by leveraging LVLM's image understanding capabilities. As shown in Figure~\ref{fig:pretrain}a, we first construct positive OCR text-image pairs $(\mathbf{h}_{\text{o}}, \mathbf{h}_{\text{d}^+})$ from raw unlabeled document images. Then, we adopt in-batch negatives to calculate the contrastive loss by InfoNCE~\cite{oord2018representation} as follows:
\begin{equation}
    \mathcal{L}_{\text{RCR}} = -\text{log} \frac{\text{exp}(\textsc{sim}(\mathbf{h}_{\text{o}}, \mathbf{h}_{\text{d}^+})/\tau
)} {\sum_{i\in\mathcal{B}} \text{exp}(\textsc{sim}(\mathbf{h}_{\text{o}}, \mathbf{h}_{\text{d}_i})/\tau
)},
\end{equation}
where $\tau$ is a temperature hyperparameter to scale the logits, and $\mathcal{B}$ represents the batch size.

\paragraph{Representation Compression via Generation (RCG).}
We propose a representation training strategy that leverages the generative capabilities of LVLMs through a customized attention mask matrix. As depicted in Figure~\ref{fig:pretrain}b, representations for the image tokens, including the \texttt{<EOS>} token, are obtained via a standard auto-regressive process. 
In contrast, for the subsequent $L$ OCR token representations, we mask the image token representations and allow only the attention of \texttt{<EOS>} token and the preceding OCR tokens. This approach facilitates pooling the image representations into \texttt{<EOS>} token. The loss function is defined as:
\begin{equation}
    \mathcal{L}_{\text{RCG}} = -\frac{1}{L}\sum_{i=1}^L\text{log}p(y_i | y_{<i}, \texttt{<EOS>}),
\end{equation}
where $y_i$ denotes the $i$-th token of the OCR.

\subsection{Supervised Fine-tuning} 
We first fine-tune the VDocRetriever with the contrastive learning objective using query-document pairs with in-batch negatives (see Figure~\ref{fig:pretrain}c). 
Then, we apply the trained VDocRetriever to search over the corpus $\mathcal{I}$ to feed the top-k documents into the VDocGenerator. Finally, we train the VDocGenerator using the next-token prediction objective. 

\section{Experiments}
\begin{table}[t!]
    \centering
        \scalebox{0.9}{
    \tabcolsep=1pt
    \small
    \begin{tabular}{lccccccc} 
        \toprule
         Dataset & Documents & \%Filtered & \#Images & \#Train\&Dev & \#Test\\ \midrule
         DocVQA~\cite{Mathew_2021_WACV} & Industry & 84.8 & 12,767 & 6,382 & -- \\
         InfoVQA~\cite{Mathew_2022_WACV} & Infographic & 61.2 & 5,485 & 9,592 & 1,048 \\
         VisualMRC~\cite{DBLP:conf/aaai/TanakaNY21} &  Webpage & 71.9 & 10,229 & 6,126 & -- \\
         ChartQA~\cite{masry-etal-2022-chartqa} & Chart & 94.0 & 20,882 & -- & 150 \\
         OpenWikiTable~\cite{kweon-etal-2023-open} & Table & 0.0 & 1,257 & 4,261 & -- \\
         DUDE~\cite{landeghem2023document} & Open & 92.3 & 27,955 & 2,135 & 496 \\
         MPMQA~\cite{zhang2023mpmqa} & Manual & 81.7 & 10,018 & 3,054 & -- \\
         SlideVQA~\cite{SlideVQA2023}\S  & Slide & 66.7 & 52,380 & -- & 760 \\
         MHDocVQA\S & Open & 9.5 & 28,550 & 9,470 & -- \\
        \bottomrule
    \end{tabular}
    }
    \caption{Datasets in OpenDocVQA. \S~denotes datasets requiring multi-hop reasoning. 
    Note that MHDocVQA was created using only the training datasets.}
    \label{tab:dataset}
\end{table}

\begin{table*}[t!]
    \centering
        \scalebox{0.86}{
    \tabcolsep=2.1pt
    \small
    \begin{tabular}{lllcccllllllll} 
        \toprule
        \multirow{2}{*}{Model} & \multirow{2}{*}{Init} & \multirow{2}{*}{Docs} & \multirow{2}{*}{Scale} & \multirow{2}{*}{\#PT} & \multirow{2}{*}{\#FT} & \multicolumn{2}{c}{\hspace{-0.5cm}ChartQA} & \multicolumn{2}{c}{\hspace{-0.5cm}SlideVQA} & \multicolumn{2}{c}{\hspace{-0.5cm}InfoVQA} & \multicolumn{2}{c}{\hspace{-0.5cm}DUDE}  \\ 
        & & & & & & Single & All & Single & All & Single & All & Single & All \\ \midrule 
        \multicolumn{13}{c}{\textit{Off-the-shelf}} \\
        BM25~\cite{robertson2009probabilistic} & -- & Text & 0 & 0 & 0 & 54.8 & 15.6 & 40.7 & 38.7 & 50.2 & 31.3 & 57.2 & 47.5 \\ 
        Contriever~\cite{izacard2021unsupervised} & BERT~\cite{DevlinCLT19} & Text & 110M & 1B & 500K & 66.9 & 59.3 & 50.8 & 46.5 & 42.5 & 21.0 & 40.6 & 29.7 \\
        E5~\cite{wang2022text} & BERT~\cite{DevlinCLT19} & Text & 110M & 270M & 1M &  74.9 & 66.3 & 53.6 & 49.6 & 49.2 & 26.9 & 45.0 & 38.9 \\ 
        GTE~\cite{li2023towards} & BERT~\cite{DevlinCLT19} & Text & 110M &788M & 3M & 72.8 & 64.7 & 55.4 & 49.1 & 51.3 & 32.5 & 42.4 & 36.0 \\ 
        E5-Mistral~\cite{wang-etal-2024-improving-text} & Mistral~\cite{jiang2023mistral7b} & Text & 7.1B & 0 & 1.85M & 72.3 & 70.0 & 63.8 & 57.6 & 60.3 & 33.9 & 52.2 & 45.2  \\ 
        NV-Embed-v2~\cite{lee2024nv} & Mistral~\cite{jiang2023mistral7b} & Text & 7.9B & 0 & 2.46M & 75.3 & 70.7 & 61.7 & 58.1 & 56.5 & 34.2 & 43.0 & 38.6\\
        CLIP~\cite{radford2021learning} & Scratch & Image & 428M & 400M & 0 & 54.6 & 38.6 & 38.1 & 29.7 & 45.3 & 20.6 & 23.2 & 17.6 \\ 
        DSE~\cite{ma2024unifying} & Phi3V~\cite{abdin2024phi} & Image & 4.2B & 0 & 5.61M & 72.7 & 68.5 & 73.0 & 67.2 & 67.4 & 49.6 & 55.5 & 47.7 \\
        VisRAG-Ret~\cite{yu2024visrag} & MiniCPM-V~\cite{yao2024minicpmvgpt4vlevelmllm} & Image & 3.4B & 0 & 240K & \textbf{87.2}* & 75.5* & 74.3* & 68.4* & 71.9* & 51.7* & 56.4 & 44.5 \\
        \midrule
        \multicolumn{13}{c}{\textit{Trained on OpenDocVQA}} \\
        Phi3~\cite{abdin2024phi} & Phi3V~\cite{abdin2024phi} & Text & 4B & 0 & 41K & 72.5 & 65.3 & 53.3 & 48.4 & 53.2* & 33.0* & 40.5* &  32.0* \\ 
        VDocRetriever\dag & Phi3V~\cite{abdin2024phi} & Image & 4.2B & 0 & 41K & 84.2$_{\textcolor{darkgreen}{+11.7}}$ & 74.8$_{\textcolor{darkgreen}{+9.5}}$ & 71.0$_{\textcolor{darkgreen}{+17.7}}$ & 65.1$_{\textcolor{darkgreen}{+16.7}}$ 
        & 66.8*$_{\textcolor{darkgreen}{+13.6}}$ & 52.8*$_{\textcolor{darkgreen}{+19.8}}$ 
        & 48.4*$_{\textcolor{darkgreen}{+7.9}}$ & 41.0*$_{\textcolor{darkgreen}{+9.0}}$\\ 
         VDocRetriever & Phi3V~\cite{abdin2024phi} & Image & 4.2B & 500K & 41K & 86.0$_{\textcolor{blue}{+1.8}}$ & \textbf{76.4}$_{\textcolor{blue}{+1.6}}$ & \textbf{77.3}$_{\textcolor{blue}{+6.3}}$ & \textbf{73.3}$_{\textcolor{blue}{+8.2}}$ &  \textbf{72.9}*$_{\textcolor{blue}{+6.1}}$ & \textbf{55.5}*$_{\textcolor{blue}{+2.7}}$ & \textbf{57.7}*$_{\textcolor{blue}{+9.3}}$ & \textbf{50.9}*$_{\textcolor{blue}{+9.9}}$ \\ 
        \bottomrule
    \end{tabular}
    }
    \caption{Retrieval results under the single- (Single) and all-pool (All) settings. * indicates performance on test data for which corresponding training samples are available. All other results represent zero-shot performance. Init, FT, and PT denote the initialization model, fine-tuning, and pre-training, respectively. 
    Performance gains in \textcolor{darkgreen}{green} and \textcolor{blue}{blue} are compared to the base LLM and VDocRetirver\dag, respectively. } 
    \label{tab:retrieval}
\end{table*}
\begin{table*}[t!]
    \centering
        \scalebox{0.86}{
    \tabcolsep=3pt
    \small
    \begin{tabular}{llllllllllll} 
        \toprule
        \multirow{2}{*}{Generator} & \multirow{2}{*}{Retriever} & \multirow{2}{*}{Docs}& \multicolumn{2}{c}{\hspace{-0.5cm}ChartQA} & \multicolumn{2}{c}{\hspace{-0.5cm}SlideVQA} & \multicolumn{2}{c}{\hspace{-0.5cm}InfoVQA} & \multicolumn{2}{c}{\hspace{-0.5cm}DUDE} \\ 
        & & & Single & All & Single & All & Single & All & Single & All & \\ \midrule 
        \multicolumn{11}{c}{\textit{Closed-book}} \\ 
        Phi3 & -- & -- & 20.0 & 20.0 & 20.3 & 20.3 & 34.9* & 34.9* & 23.1* & 23.1* \\ 
        \midrule
        \multicolumn{11}{c}{\textit{Text-based RAG}} \\
        Phi3 & Phi3 & Text & 28.0 & 28.0 & 28.6 & 28.0 & 40.5* & 39.1* & 40.1* & 35.7* \\
        \hgrow Phi3 & Gold & Text & 36.6 & 36.6 & 27.8 & 27.8 & 45.6* & 45.6* & 55.9* & 55.9* \\ 
        \midrule
        \multicolumn{11}{c}{\textit{VDocRAG (Ours)}} \\
        VDocGenerator & VDocRetriever & Image &  \textbf{52.0}$_{\textcolor{darkgreen}{+24.0}}$ & \textbf{48.0}$_{\textcolor{darkgreen}{+20.0}}$ &  \textbf{44.2}$_{\textcolor{darkgreen}{+15.6}}$ & \textbf{42.0}$_{\textcolor{darkgreen}{+14.0}}$ & \textbf{56.2}*$_{\textcolor{darkgreen}{+15.7}}$ & \textbf{49.2}*$_{\textcolor{darkgreen}{+10.1}}$ &\textbf{48.5}*$_{\textcolor{darkgreen}{+8.4}}$ &\textbf{44.0}*$_{\textcolor{darkgreen}{+8.3}}$  \\
        \hgrow VDocGenerator & Gold & Image & 74.0 & 74.0 & 56.4 & 56.4 & 64.6* & 64.6* & 66.4* & 66.4* \\ 
        \bottomrule
    \end{tabular}
    }
    \caption{DocumentVQA results. All models are fine-tuned on OpenDocVQA. The results marked with * denote performance on unseen test samples, and the other results represent zero-shot performance. The performance gain in \textcolor{darkgreen}{green} is compared to the text-based RAG that has the same base LLM. Gold knows the ground-truth documents. Models answer the question based on the top three retrieval results.}
    \label{tab:rag}
\end{table*}
\subsection{Experimental Setup}

\paragraph{Pre-training dataset.}
For pre-training, we gathered 500k samples containing document image and OCR text pairs filtered from the DocStruct4M~\cite{hu2024mplug}. We excluded any images that appeared in the test set to avoid data contamination.

\paragraph{Fine-tuning and evaluation datasets.}
We evaluated our models in both zero-shot and supervised settings. The zero-shot evaluation assessed the models' generalization capabilities on unseen datasets, while the supervised evaluation measured performance when training samples were available. As shown in Table~\ref{tab:dataset}, we trained our models on seven datasets and evaluated them on four datasets, including ChartQA and SlideVQA in the zero-shot setting, and InfoVQA and DUDE in the supervised setting. 

\paragraph{Implementation details.}
We initialized VDocRAG with Phi3V~\cite{abdin2024phi}, a state-of-the-art LVLM trained on high-resolution images and multi-image data. The parameters of VDocRetriever and VDocGenerator were not shared. We employed LoRA~\cite{hu2021lora} with LLM while keeping other parameters frozen during training. We trained VDocRAG for one epoch on eight A100-80G GPUs with AdamW~\cite{loshchilov2017decoupled} optimizer and FlashAttention~\cite{dao2022flashattention}, using batch sizes of 16 for pre-training and 64 for fine-tuning. We set the temperature $\tau$ to 0.01. We applied Tesseract~\cite{smith2007overview} to extract OCR text in images. By default, we used the top three documents obtained from VDocRetirver. 

\paragraph{Retrieval baselines.}
We compared VDocRetriever with two categories of retrievers. The first category includes off-the-shelf text retrieval models on extracted text and image retrieval models. These consist of \textbf{BM25}~\cite{robertson2009probabilistic}, a lexical matching model; \textbf{Contriver}~\cite{izacard2021unsupervised}, \textbf{E5}~\cite{wang2022text}, and \textbf{GTE}~\cite{li2023towards}, which are popular strong text embedding models based on BERT~\cite{DevlinCLT19}; \textbf{E5-Mistral}~\cite{wang-etal-2024-improving-text} and \textbf{NV-Embed-v2}~\cite{lee2024nv}, which are state-of-the-art LLM-based embedding models; \textbf{CLIP}~\cite{radford2021learning}, a dual-encoder vision-language model; \textbf{DSE}~\cite{ma2024unifying} and \textbf{VisRAG-Ret}~\cite{yu2024visrag}, which are state-of-the-art visual document retrieval models. The second category includes fine-tuned models trained on OpenDocVQA. To verify the effectiveness of encoding documents through images, 
we fine-tuned the LLM in VDocRetriever (\textbf{Phi3}~\cite{abdin2024phi}) using extracted text to represent documents. Additionally, we included a variant of VDocRetriever without pre-training (\textbf{VDocRetriever\dag}).

\paragraph{QA baselines.}
We compared VDocRAG against \textbf{closed-book} and \textbf{text-based RAG} models. These baselines used the same model initialization as VDocRAG but fine-tuned only the LLM (Phi3). The closed-book model received only the question as input, while the text-based RAG used the top three documents retrieved by the Phi3 retriever. Moreover, we assessed possible upper-bound performance by testing generation with ground-truth (Gold) documents.

\paragraph{Evaluation metrics.}
We evaluated retrieval performance using \textbf{nDCG@5}, a widely used metric in information retrieval~\cite{kamalloo2023resources,faysse2024colpali}. For the DocumentVQA task, we followed the evaluation protocol of each dataset, we used \textbf{ANLS}~\cite{BitenTMBRJVK19} for InfoVQA and DUDE, \textbf{Relaxed Accuracy}~\cite{masry-etal-2022-chartqa} for ChartQA, \textbf{F1} for SlideVQA as evaluation metrics.

\subsection{Retrieval Results}
Table~\ref{tab:retrieval} shows that VDocRetriever\dag~achieved significantly higher retrieval performance than the text-based Phi3 retriever on all datasets under the same conditions. This indicates that our model can effectively encode documents in image format for retrieval tasks. Furthermore, VDocRetriever exhibits superior zero-shot generalization on unseen datasets, ChartQA and SlideVQA, outperforming both off-the-shelf text retrievers and state-of-the-art visual document retrieval models. Notably, DSE was initialized with the same LVLM as ours and fine-tuned on 13.7 times more data. This highlights that our pre-training strategy and the OpenDocVQA dataset offer unique advantages that are not adequately addressed by existing approaches.

\subsection{Retrieval-Augmented Generation Results}
Table~\ref{tab:rag} shows that VDocRAG significantly outperformed both the closed-book LLM and the text-based RAG on the DocumentVQA task, even when all models were the same initialization. Additionally, when the retrieval results were fixed to ground-truth (Gold) documents, VDocRAG demonstrated superior performance to text-based RAG. This underscores the importance of visual cues in extracting answers from documents and suggests that VDocGenerator has a higher upper-bound performance. Both text-based RAG and VDocRAG exhibited substantial improvements when provided with ground-truth documents, highlighting potential areas for enhancing retrieval accuracy and improving the generator's robustness to retrieval noise.

\begin{table}[t!]
    \centering
        \scalebox{0.95}{
    \tabcolsep=3pt
    \small
    \begin{tabular}{lll} 
        \toprule
        Model & SlideVQA & InfoVQA \\ \midrule
        VDocRetriever & \textbf{77.3}  & \textbf{72.9} \\ \hdashline
        w/o RCR & 75.9$_{\textcolor{red}{-1.4}}$ & 71.1$_{\textcolor{red}{-1.8}}$ \\
        w/o RCG  & 71.7$_{\textcolor{red}{-5.6}}$ & 68.8$_{\textcolor{red}{-4.1}}$ \\ 
        w/o RCG \& RCR & 71.0$_{\textcolor{red}{-6.3}}$ & 66.8$_{\textcolor{red}{-6.1}}$  \\ 
        w/o LLM \& Projector ($\hookrightarrow$ CLIP encoders) & 43.7$_{\textcolor{red}{-33.6}}$ & 37.9$_{\textcolor{red}{-35.0}}$\\ 
        \bottomrule
    \end{tabular}
    }
    \caption{Ablation study of our pre-training tasks and model architecture in the retrieval task under the single-pool setting.}
    \label{tab:ablation}
\end{table}

\begin{table}[t!]
    \centering
        \scalebox{0.9}{
    \tabcolsep=2.5pt
    \small
    \begin{tabular}{lllll} 
        \toprule
        \multirow{2}{*}{Model} & \multicolumn{2}{c}{Retrieval} & \multicolumn{2}{c}{QA} \\         
         & {SlideVQA} & {InfoVQA} & {SlideVQA} & {InfoVQA} \\ 
        \midrule
        VDocRAG & \textbf{77.3} & \textbf{72.9} & \textbf{44.2} & \textbf{56.2} \\ \hdashline
        w/o MHDocVQA & 75.0$_{\textcolor{red}{-2.3}}$ & 71.4$_{\textcolor{red}{-1.5}}$ & 43.4$_{\textcolor{red}{-0.8}}$ & 53.8$_{\textcolor{red}{-2.4}}$ \\
        w/o except MHDocVQA & 68.8$_{\textcolor{red}{-8.5}}$ & 61.7$_{\textcolor{red}{-11.2}}$ & 41.1$_{\textcolor{red}{-3.1}}$ & 44.0$_{\textcolor{red}{-12.2}}$ \\ 
        \bottomrule
    \end{tabular}
    }
    \caption{Ablation study of our dataset in retrieval and QA tasks under the single-pool setting.}
    \label{tab:ablation_dataset}
\end{table}

\subsection{Analysis}
\begin{figure}[t!]
    \centering
    \includegraphics[width=.48\textwidth]{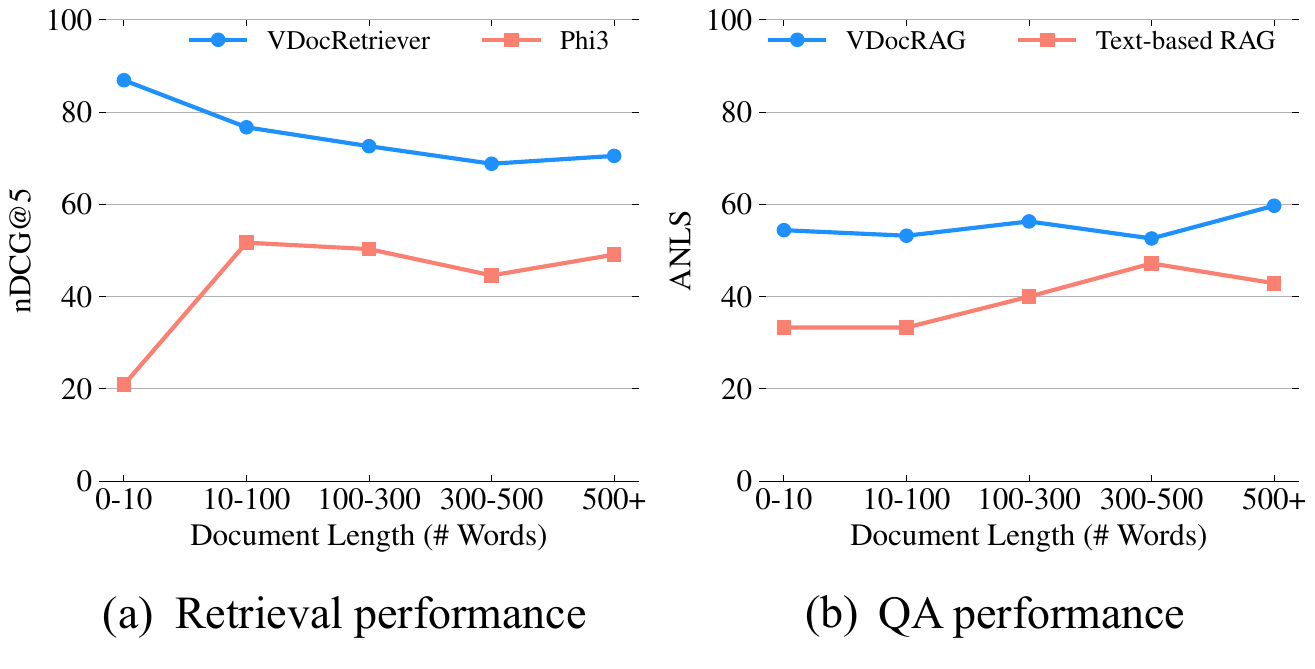}
    \caption{Performance under different document lengths on InfoVQA (single-pool setting).}
    \label{fig:text_dense}
\end{figure}

\paragraph{Can our pre-training tasks be beneficial?}
Table~\ref{tab:ablation} shows that VDocRetriever outperformed the model without pre-training. Removing each pre-training task or both RCG and RCR tasks decreased performance, indicating that both tasks contribute complementarily. These validate that our pre-training effectively learns to compress image features while aligning them with textual contents in images.

\paragraph{Does LLM help understanding document images?}
Table~\ref{tab:ablation} shows that retrieval performance dropped substantially when the LLM block was removed, leaving only the CLIP text/vision encoder, even with the same visual transformer backbone. This suggests that LLM can capture finer-grained visual details and enhance semantic understanding.

\begin{table}[t!]
    \centering
        \scalebox{0.95}{
    \tabcolsep=2.5pt
    \small
    \begin{tabular}{lcccc} 
        \toprule
        & \multicolumn{2}{c}{Retrieval} & QA & \\
        Model & OCR & Encoding & Generation & Total \\ \midrule
        Text-based RAG$_{\text{Phi3}}$ & 590.0 & 70.7 & 422.7 & 1083.4 \\
        VDocRAG & -- & 204.4 & 789.7 & 994.1 \\
        \bottomrule
    \end{tabular}
    }
    \caption{Efficiency analysis on InfoVQA. The average time (ms) to encode a single document or generate a single answer is measured on a single A100 GPU.}
    \label{tab:efficiency}
\end{table}

\begin{table}[t!]
    \centering
        \scalebox{0.93}{
    \tabcolsep=2.5pt
    \small
    \begin{tabular}{lcccc} 
        \toprule
        \multirow{2}{*}{Model} & \multicolumn{2}{c}{Retrieval} & \multicolumn{2}{c}{QA} \\         
         & {SlideVQA} & {InfoVQA} & {SlideVQA} & {InfoVQA} \\ 
        \midrule
        Text-based RAG$_{\text{LLama3}}$ & 60.1 & 61.8 & 37.8 & 49.5 \\ 
        VDocRAG$_{\text{Idefics3}}$ & \textbf{73.4} & \textbf{72.5}& \textbf{48.9} & \textbf{59.9}\\ 
        \hspace{0.3cm}w/o Pre-train & 70.3 & 69.8 & 47.2 & 59.6 \\
        \bottomrule
    \end{tabular}
    }
    \caption{Analysis with different LVLM (Idefics3) in retrieval and QA tasks under the single-pool setting.}
    \label{tab:idefics3}
\end{table}
\paragraph{Does our dataset improve the performance?}
Table~\ref{tab:ablation_dataset} shows that removing MHDocVQA caused a performance decrease, indicating that MHDocVQA requires distinct reasoning skills compared to other collected datasets in OpenDocVQA. Additionally, excluding all OpenDocVQA datasets except MHDocVQA led to a significant performance drop. This confirms that our collected datasets effectively supplement the missing capabilities of LVLM in document retrieval and understanding.

\paragraph{How well does VDocRAG perform under different document lengths?}
Figure~\ref{fig:text_dense} shows that VDocRAG consistently outperforms text-based RAG, indicating that VDocRAG can better understand documents through visual information. In general, we observed that the VDocRAG's relative performance over text-based RAG is larger for images with 0-10 words (+66.0 in retrieval, +21.1 in QA) than for those with 500+ words (+28.4 in retrieval, +16.7 in QA).

\begin{figure*}[t!]
    \centering
\includegraphics[width=1.0\textwidth]{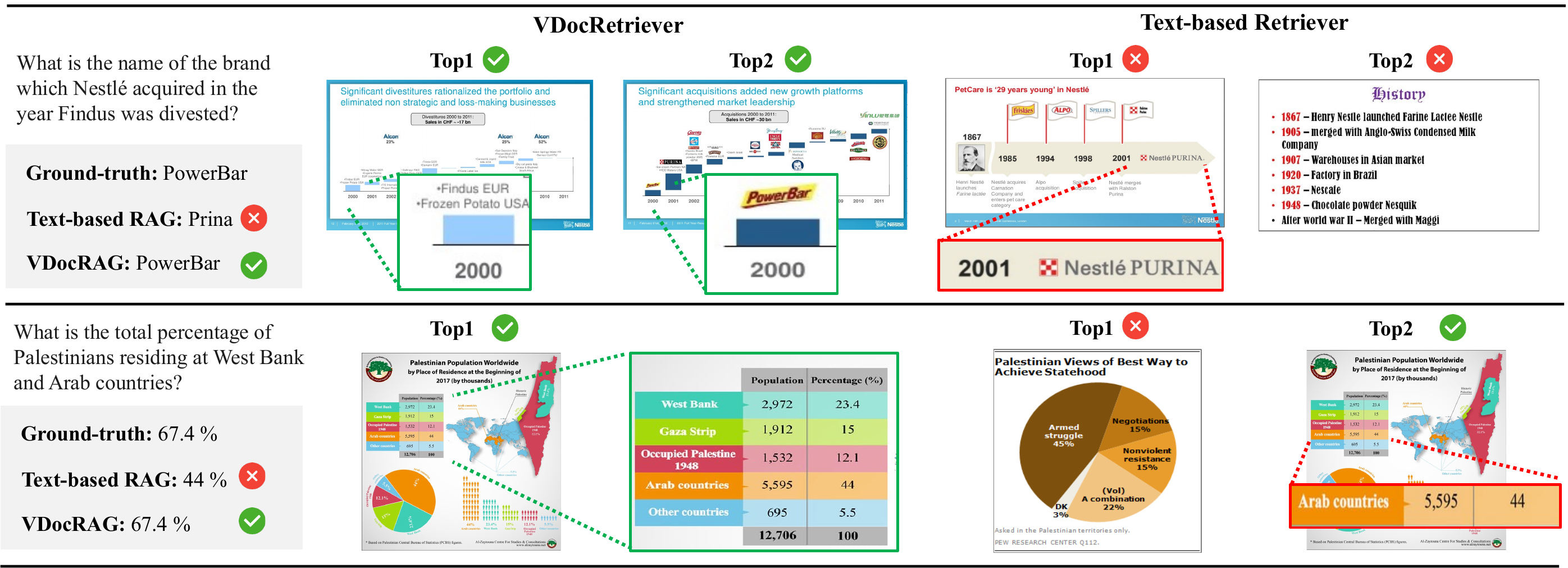}
    \caption{Qualitative results of VDocRAG compared to text-based RAG.}
    \label{fig:output}
\end{figure*}

\paragraph{Is VDocRAG more efficient than text-based RAG?}
Table~\ref{tab:efficiency} shows that VDocRAG is more efficient than text-based RAG. Especially, VDocRAG requires 69\% less inference time to retrieve documents than text-based RAG. Although VDocRetriever takes more time for document encoding and generation, it eliminates the time-consuming OCR processing necessary for text-based RAG.

\paragraph{Can our method apply different LVLMs?}
To investigate the impact of different LVLMs on VDocRAG, we replaced Phi3V with Idefics3~\cite{laurenccon2024building}, a state-of-the-art LVLM that uses Llama3-8B~\cite{dubey2024llama} as its backbone LLM. As observed in Table~\ref{tab:idefics3}, the performance trend was consistent with that of Phi3V, highlighting the versatility and broad applicability of our method.

\paragraph{Qualitative results.} Figure~\ref{fig:output} illustrates the performance of our model through qualitative examples. In the top example, VDocRAG demonstrates strong performance on a question requiring multi-hop reasoning and graph understanding across multi-page slides. In the bottom example, VDocRAG also performs better on a question that requires parsing on the table with cells spanning multiple rows and columns. In contrast, text-based RAG depends solely on OCR text information, leading to a superficial understanding of the text and incorrect predictions.

\begin{figure}[t!]
    \centering
    \includegraphics[width=.48\textwidth]{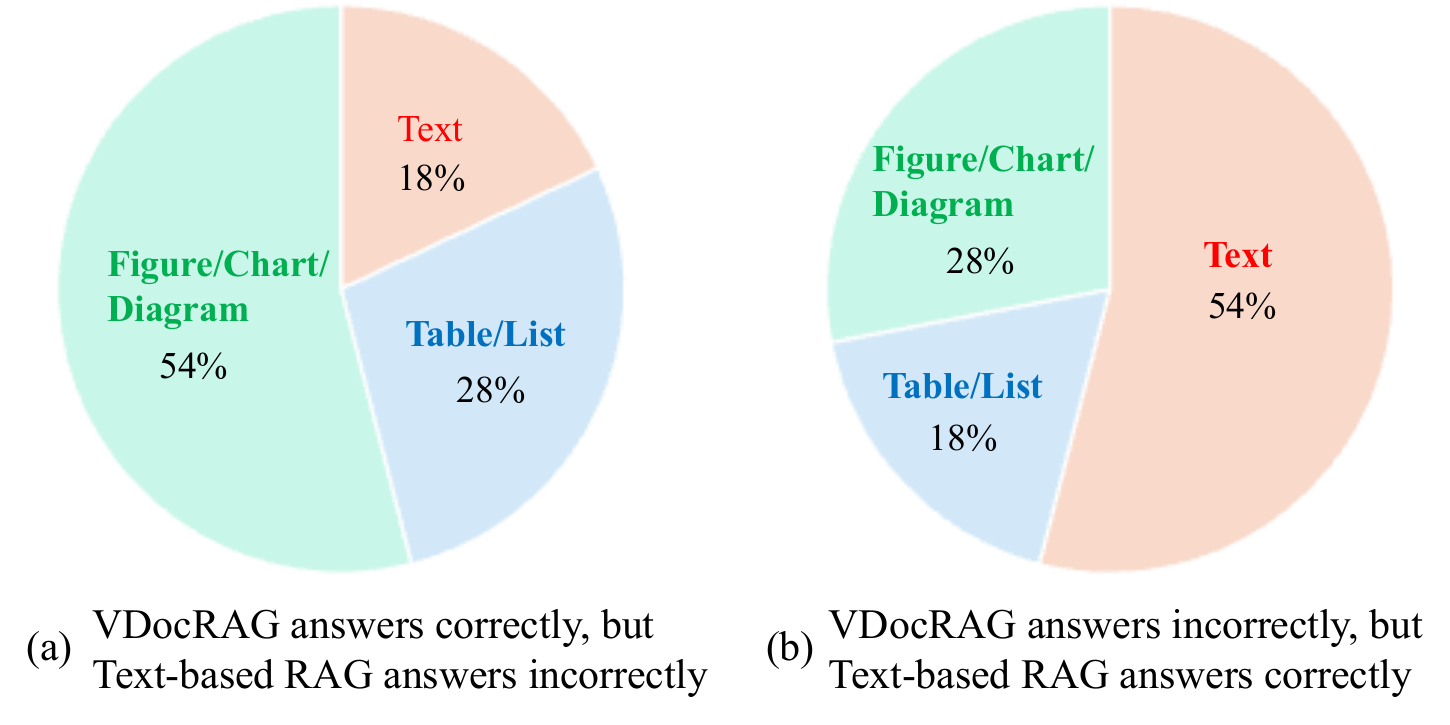}
    \caption{Root causes of correct and incorrect predictions.}
    \label{fig:error}
\end{figure}

\paragraph{Human evaluation.} To better understand the prediction differences between VDocRAG and text-based RAG, we manually analyzed the generated outputs by identifying the root causes of 50 correct and 50 incorrect predictions, randomly sampled from test samples. Figure~\ref{fig:error}a shows that VDocRAG significantly enhances the understanding of visual data (e.g., charts). Conversely, Figure~\ref{fig:error}b reveals that VDocRAG encounters challenges with text-heavy documents (e.g., books), primarily due to the OCR capabilities. We observed that text-based RAG correctly answers questions when visual data includes long titles or subtitles, which have a high textual overlap with the question. These observations are in line with the results shown in Figure~\ref{fig:text_dense}.

\section{Conclusion}
We introduced a new RAG framework, VDocRAG, which can directly understand various real-world documents. We enhanced VDocRAG with two key contributions: (1) pre-training tasks capable of learning image representation efficiently by leveraging the powerful capabilities of LVLMs, and (2) OpenDocVQA, the first unified open-domain DocumentVQA dataset that encompasses a wide range of visually-rich documents. Our holistic evaluations on four datasets show that VDocRAG significantly outperformed conventional text-based RAG, shedding light on the development of an effective RAG over real-world documents. 

\paragraph{Limitations.}
While we focused on pre-training to align images and OCR data for document retrieval, leveraging caption data instead of OCR data offers the potential for retrieving images that do not contain text. Moreover, this study did not address reducing the computational cost of creating search indexes for extensive image collections. We plan to reduce the cost of VDocRAG using more efficient techniques. Lastly, joint training of QA and retrieval components simultaneously further optimizes their interactions.

{
    \small
    \bibliographystyle{ieeenat_fullname}
    \bibliography{main}
}

\input{X_suppl}

\end{document}

%% file: X_suppl.tex
\renewcommand\thefigure{\Alph{section}.\arabic{figure}}
\renewcommand\thetable{\Alph{section}.\arabic{table}} 

\setcounter{figure}{0}
\setcounter{table}{0}

\newcommand\beginsupplement{%
        \setcounter{table}{0}
        \renewcommand{\thetable}{\Alph{table}}%
        \setcounter{figure}{0}
        \renewcommand{\thefigure}{\Alph{figure}}%

     }
\beginsupplement

\clearpage
\setcounter{page}{1}
\maketitlesupplementary

\appendix

\section{OpenDocVQA Details}
\begin{table}[t!]
    \centering
        \scalebox{1.0}{
    \tabcolsep=3pt
    \small
    \begin{tabular}{lc} 
        \toprule
        Statistics & Number \\ \midrule
        Total Images & 206,267 \\ \midrule
        Total Questions & 43,474 \\ 
        - Single-Hop Questions & 33,244 (76.5\%) \\ 
        - Multi-Hop Questions & 10,230 (23.5\%) \\ 
        - Extractive Answer & 19,797 (45.5\%) \\
        - Abstractive Answer & 23,677 (54.5\%) \\ \midrule
        QA Source Datasets & 9 \\
        - Existing DocumentVQA Datasets & 7 \\
        - Existing TableQA Datasets & 1 \\
        - Our Newly Created Datasets & 1 \\ \midrule
        Maximum Question Length & 58 \\
        Maximum Answer Length & 130 \\
        Average Question Length & 13.7 \\
        Average Answer Length & 3.7 \\
        \bottomrule
    \end{tabular}
    }
    \caption{Main statistics in OpenDocVQA.}
    \label{tab:stat}
\end{table}

\paragraph{Dataset Statistics.}
The main statistics of OpenDocVQA are presented in Table~\ref{tab:stat}. There are two types of questions: single-hop (45.5\%) and multi-hop (23.5\%). Answers to questions are categorized as extractive (45.5\%) and abstractive (54.5\%) types. OpenDocVQA consists of nine open-domain DocumentVQA datasets, including a newly created MHDocVQA dataset to address multi-hop questions over multiple documents, and collected and filtered QA datasets as follows. 
\begin{itemize}
    \item \textbf{DocVQA}~\cite{Mathew_2021_WACV} includes industry document images collected from the UCSF Industry Document Library.
    \item \textbf{InfoVQA}~\cite{Mathew_2022_WACV} includes infographics downloaded from the Internet for the search query ``infographics".
    \item \textbf{VisualMRC}~\cite{DBLP:conf/aaai/TanakaNY21} is a visual machine reading comprehension on webpage screenshot images.
    \item \textbf{ChartQA}~\cite{masry-etal-2022-chartqa} is a chart understanding dataset with human-written and machine-generated questions focusing on visual and logical reasoning.
    \item \textbf{OpenWikiTable}~\cite{kweon-etal-2023-open} is an open-domain question answering over tables. We took screenshot images of the tables, converting them into images with complex text layouts to handle visually-rich table data.
    \item \textbf{DUDE}~\cite{landeghem2023document} is a multi-page, multi-domain, and multi-industry QA dataset that requires processing long documents and understanding different types of documents.
    \item \textbf{MPMQA}~\cite{zhang2023mpmqa} requires comprehending multimodal content in an entire product manual and answering questions.
    \item \textbf{SlideVQA}~\cite{SlideVQA2023} requires multi-hop reasoning over multiple slide images containing various text formats, layouts, and visual content such as plots and charts.
\end{itemize}

\begin{figure}[!t]
\centering
\subfloat[Word cloud of questions.]{\includegraphics[clip, width=.335\textwidth]{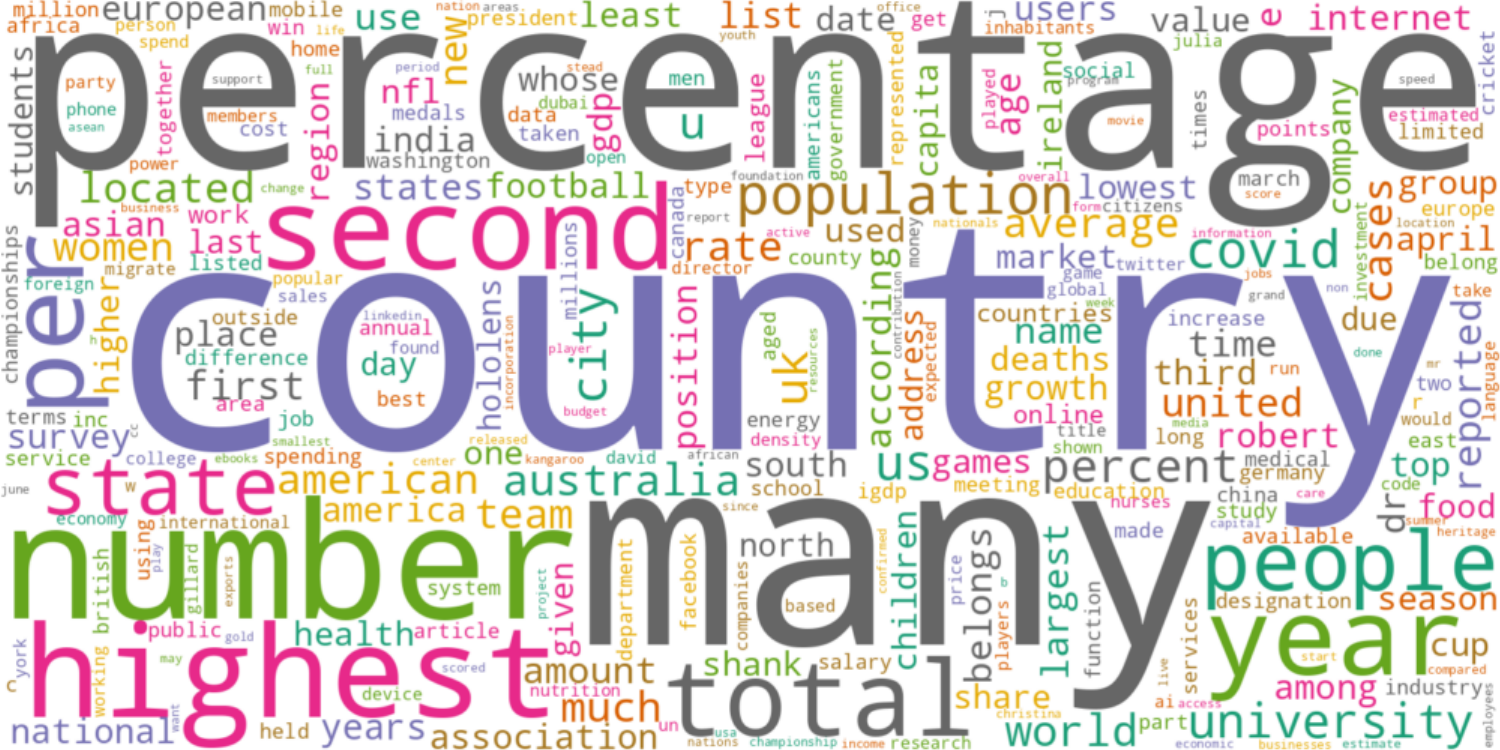}
\label{fig:label-A}}
\\
\subfloat[Word cloud of answers.]{\includegraphics[clip, width=.335\textwidth]{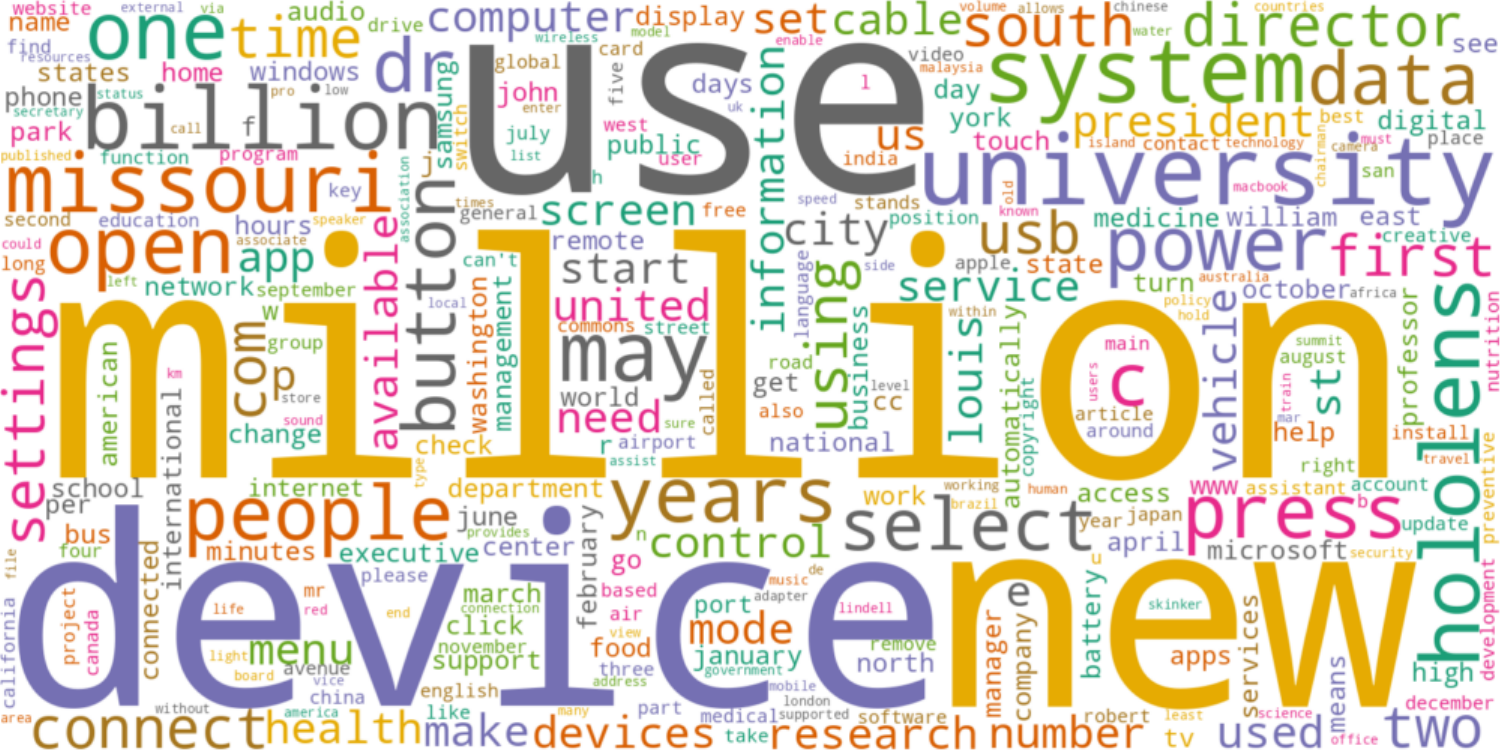}
\label{fig:label-B}}
\caption{Word cloud distributions of question and answer texts.}
\label{fig:cloud}
\end{figure}

\begin{figure}
    \centering
    \includegraphics[width=.36\textwidth]{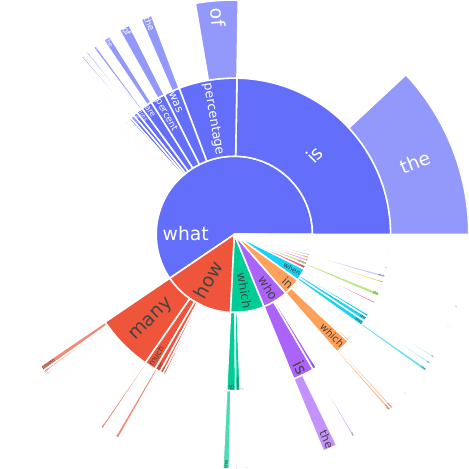}
    \caption{Distribution of first three words of the question.}
    \label{fig:sunburst}
\end{figure} 

Figure~\ref{fig:cloud} presents word clouds of the most frequently appeared words in the question and answer texts, illustrating that OpenDocVQA covers a wide range of topics and words. This observation is further supported by Figure~\ref{fig:sunburst}, which is a sunburst of the first three words of the questions.

\begin{table*}[t!]
\centering
\begin{tcolorbox}[fontupper=\footnotesize, title=\small Multi-hop Question Generation Prompt]
\begin{verbatim}
EXAMPLE1:
question1: In which country is the GWP smallest?
answer1: Denmark
question2: What is the staple diet of Denmark?
answer2: Fish, cheese
combined question: What is the staple diet of the country where the GWP is the smallest?

EXAMPLE2:
question1: To which League does Chicago Cubs belong?
answer1: mlb
question2: What is the average MLB team value?
answer2: $1.5b
combined question: What is the average the league where Chicago Cubs belongs to team value?

EXAMPLE3
question1: Which is the capital city of Germany?
answer1: Berlin
question2: What year did Berlin host the OKFestival?
answer2: It's 2014.
combined question: What year did the capital city of Germany host the OKFestival?

Based on the above 3 examples, provide a combined question for the following case, 
such that the answer to the combined question is the same as the answer2:
question1: {single-hop question}
answer1: {single-hop answer}
question2: {single-hop question}
answer2: {single-hop answer}
combined question: 
\end{verbatim}
\end{tcolorbox}
\caption{Multi-hop question generation prompt. ``\{single-hop question\}" and ``\{single-hop answer\}" are placeholders of two single-hop questions.}
\label{tab:combine_prompt}
\end{table*}

\begin{table*}[t!]
\centering
\begin{tcolorbox}[fontupper=\footnotesize, title=\small Multi-hop Question Filtering Prompt]
\begin{verbatim}
question1: {single-hop question}
answer1: {single-hop answer}
question2: {single-hop question}
answer2: {single-hop answer}

Based on the questions and answers above, please answer the following question shortly. 
If the answer is not identified, the answer is 'None': {multi-hop question}
\end{verbatim}
\end{tcolorbox}
\caption{Multi-hop question filtering prompt. ``\{single-hop question\}" and ``\{single-hop answer\}" are placeholders of two single-hop questions. ``\{multi-hop question\}" denotes the generated multi-hop questions.}
\label{tab:fileter_prompt}
\end{table*}

\begin{table*}[t!]
    \centering
        \scalebox{0.99}{
    \tabcolsep=3pt
    \small
    \begin{tabular}{ll} 
        \toprule
        Dataset & Task Description \\ \midrule
        DocVQA & You have to find an industry document that answers my question. \\
        InfoVQA & Given a question, retrieve an infographic to answer the question. \\
        VisualMRC & I'm looking for a screenshot image that answers the question. \\
        ChartQA & Given a user query, retrieve a chart image that answers the query. \\
        OpenWikiTable & Given a user query, retrieve a table image for answering the question. \\
        DUDE & You need to retrieve evidence from a PDF page to address the question.\\
        MPMQA & I want to know the answer to the question. Can you find evidence from manual pages?\\
        SlideVQA & Given a question, retrieve a slide image to answer the question.\\
        MHDocVQA & Given a multihop-question, retrieve multiple pages that can help answer the question.\\
        \bottomrule
    \end{tabular}
    }
    \caption{Instructions in the visual document retrieval task.}
    \label{tab:instruction}
\end{table*}

\begin{table}[t!]
    \centering
        \scalebox{0.88}{
    \tabcolsep=3pt
    \small
    \begin{tabular}{ll} 
        \toprule
        Model & Model Checkpoint \\ \midrule
        Contriever & \texttt{facebook/contriever-msmarco}  \\ 
        E5 & \texttt{intfloat/e5-base-v2} \\
        GTE & \texttt{thenlper/gte-base} \\
        E5-Mistral & \texttt{intfloat/e5-mistral-7b-instruct} \\
        NV-Embed-v2 & \texttt{nvidia/NV-Embed-v2} \\
        CLIP & \texttt{openai/clip-vit-large-patch14-336} \\
        DSE & \texttt{Tevatron/dse-phi3-docmatix-v1} \\
        VisRAG-Ret & \texttt{openbmb/VisRAG-Ret} \\
        Phi3V & \texttt{microsoft/Phi-3-vision-128k-instruct} \\
        Idefics3 & \texttt{HuggingFaceM4/Idefics3-8B-Llama3} \\
        \bottomrule
    \end{tabular}
    }
    \caption{Model checkpoints stored on HuggingFace.}
    \label{tab:checkpoints}
\end{table}

\begin{table}[t!]
    \centering
        \scalebox{0.90}{
    \tabcolsep=3pt
    \small
    \begin{tabular}{lc} 
        \toprule
        Hyperparameters & Value \\ \midrule
        Learning Rate & 1e-4 \\ 
        Gradient Accumulation & 4 \\
        Adam W $\beta_1$ & 0.9 \\
        Adam W $\beta_2$ & 0.999 \\
        LoRA Attention Dimension r & 8 \\
        LoRA Scaling Alpha & 64 \\
        LoRA Dropout & 0.1 \\
        LoRA Target & *\_proj \\
        BF16  & True \\
        \bottomrule
    \end{tabular}
    }
    \caption{Hyperparameters used for pre-training and fine-tuning.}
    \label{tab:hyperparams}
\end{table}

\paragraph{Filtering DocumentVQA datasets.} We applied the following five heuristic rules to automatically filter out likely context-dependent questions:

\begin{itemize}
    \item The question has one or more demonstrative pronouns, including ``this", ``these", and ``those".
    \item The question has one or more personal pronouns, including  ``she", ``he", ``her", ``his", and ``him". 
    \item The question has one or more specific keywords, including ``the document" and ``mention".
    \item The question does not contain entities except for numbers. 
    \item The question is shorter than six words. 
\end{itemize}
Any samples matching at least one of these rules were removed from our dataset. After applying the rules, we manually reviewed all the questions to ensure context-independence, guided by the instruction: ``\textit{When you see the question without a given document, can you find a unique document in the corpus to provide a unique answer?}". To validate our review, we randomly sampled 50 questions with their gold and top-5 retrieved documents (from VDocRetriever) and found no ambiguous cases, confirming the high quality of our process.

\paragraph{Prompts for creating multi-hop questions.} Table~\ref{tab:combine_prompt} shows the prompt for combining two single-hop questions to generate multi-hop questions. Moreover, Table~\ref{tab:fileter_prompt} shows the prompt for filtering the generated multi-hop questions.

\section{Experimental Details}

\paragraph{Instruction templates.}
Following a standard LLM-based retrieval training and evaluation strategy~\cite{wang-etal-2024-improving-text}, we applied natural language instruction templates to the original question for the visual document retrieval task: 
\begin{equation*}
    \text{Instruct: \{task description\} \textbackslash n Query: \{question\}},
\end{equation*}
where ``\{task description\}" is a placeholder for a one-sentence task description as shown in Table~\ref{tab:instruction}. Note that the instruction format was applied to only LLM-based retrievers, including E5-Mistral~\cite{wang-etal-2024-improving-text}, NV-Embed-v2~\cite{lee2024nv}, DSE~\cite{ma2024unifying}, Phi3~\cite{abdin2024phi}, and VDocRetriever. Our preliminary experiments observed that using the instruction during both training and evaluation improved the performance of LLM-based retrievers. However, applying the same instruction format to non-LLM-based retrievers, such as Contriever~\cite{izacard2021unsupervised}, resulted in a performance decline due to lacking instruction-following capabilities. Furthermore, we appended an instruction regarding the desired output format for the DocumentVQA task:
\begin{equation*}
\text{\textbackslash n Answer briefly.}
\end{equation*}

\paragraph{Model checkpoints}
Table~\ref{tab:checkpoints} shows model initialization checkpoints stored on HuggingFace~\footnote{\url{https://huggingface.co}}. 

\paragraph{Model hyperparameters}
Table~\ref{tab:hyperparams} lists hyperparameters in pre-training and fine-tuning used for our models. 

\begin{table}[t!]
    \centering
        \scalebox{0.95}{
    \tabcolsep=2.5pt
    \small
    \begin{tabular}{lcccc} 
        \toprule
        Max Image & \multicolumn{2}{c}{Retrieval} & \multicolumn{2}{c}{QA} \\        
        Resolution & {nDCG@5} & {Encoding Time} & {ANLS} & {Generation Time} \\ 
        \midrule
        336$\times$336 &  28.7 & 85.0 & 37.2 & 394.5 \\ 
        672$\times$672 &  72.8 & 106.4 & 42.7 & 490.9 \\
        1344$\times$1344 & 72.9 & 204.4 & 56.2 & 789.7 \\ 
        \bottomrule
    \end{tabular}
    }
    \caption{Impact of image resolution on InfoVQA under the single-pool setting. Average time (ms) to encode a single document or generate a single answer is measured on a single A100 GPU. 
    }
    \label{tab:resolution}
\end{table}


\begin{figure}[t!]
    \centering
    \includegraphics[width=.45\textwidth]{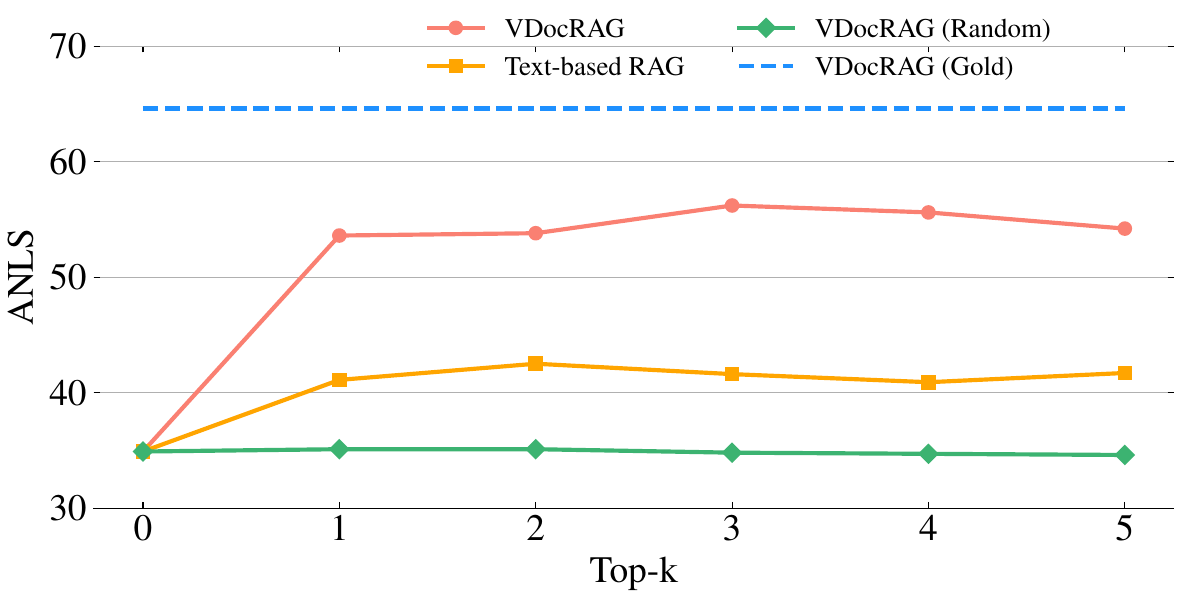}
    \caption{QA performance with various top-k on InfoVQA under the single-pool setting. () denotes document sources.}
    \label{fig:k_retrives}
\end{figure}

\section{Additional Experimental Analysis}
\paragraph{How does image resolution impact performance?}
Table~\ref{tab:resolution} shows that increasing image resolution improved the model’s capability to understand and encode the document; however, it also significantly increased the inference time for both retrieval and QA tasks. Moreover, the performance in the QA task exhibited greater sensitivity to image resolution compared to the retrieval task, indicating that the QA task demands more detailed visual understanding.

\paragraph{How many retrieved documents to augment?}
Figure~\ref{fig:k_retrives} shows that incorporating three documents yielded the best results in VDocRAG. While adding a few documents may include helpful contexts, adding more low-ranked or randomly sampled documents introduces noise and deteriorates generation due to the imperfections of retrievers.

\begin{figure*}
    \centering
    \includegraphics[width=1.0\linewidth]{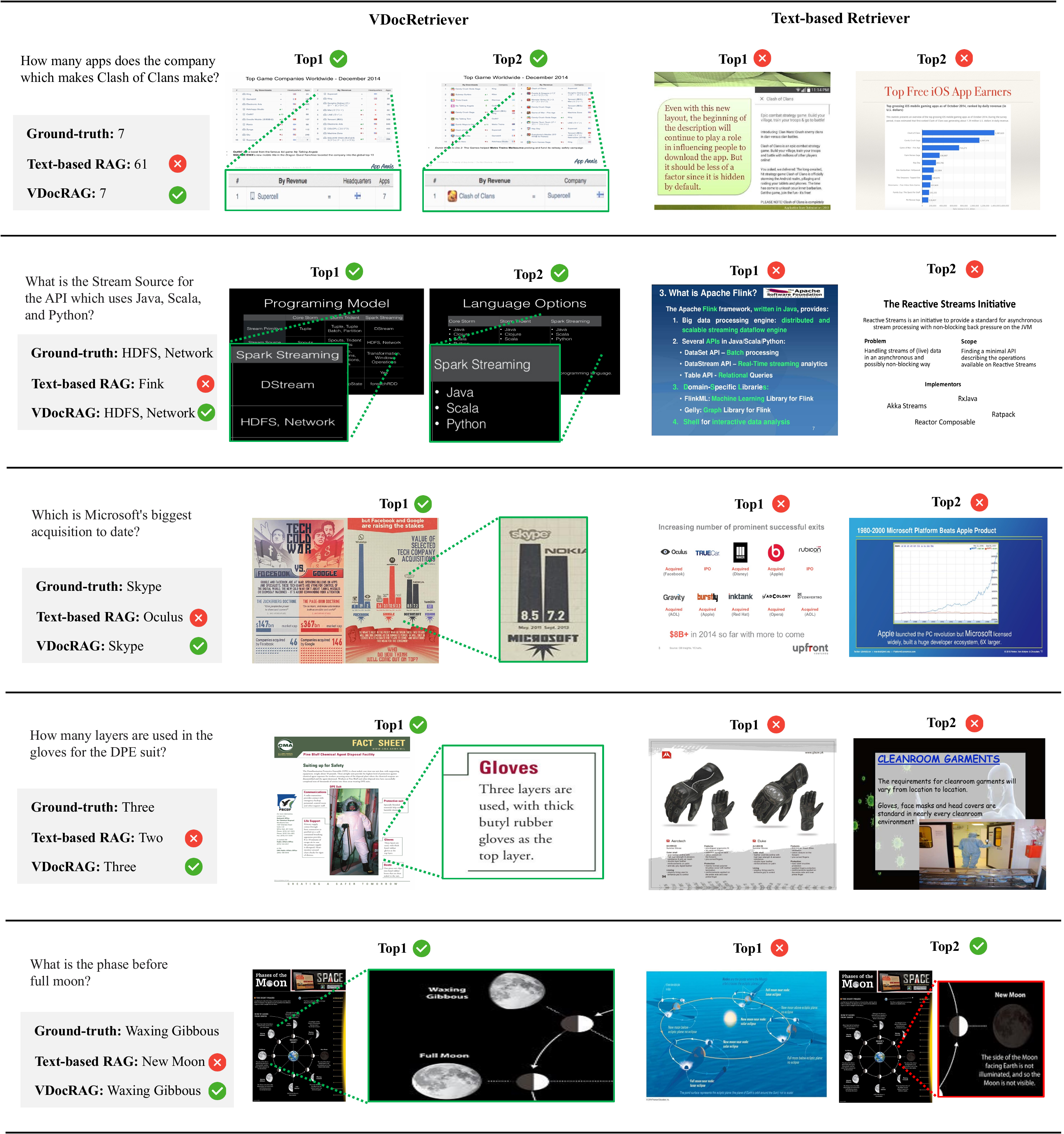}
    \caption{Additional qualitative results of VDocRAG compared to Text-based RAG.}
    \label{fig:appendix_outputs}
\end{figure*}

\paragraph{Additional qualitative results.}
Figure~\ref{fig:appendix_outputs} shows qualitative results of VDocRAG compared to text-based RAG. VDocRAG demonstrates significant performance advantages in understanding layouts and visual content, such as tables, charts, figures, and diagrams. These findings highlight the critical role of representing documents as images to improve the performance of the RAG framework.

%% file: main.bbl
\begin{thebibliography}{71}
\providecommand{\natexlab}[1]{#1}
\providecommand{\url}[1]{\texttt{#1}}
\expandafter\ifx\csname urlstyle\endcsname\relax
  \providecommand{\doi}[1]{doi: #1}\else
  \providecommand{\doi}{doi: \begingroup \urlstyle{rm}\Url}\fi

\bibitem[Abdin et~al.(2024)Abdin, Jacobs, Awan, Aneja, Awadallah, Awadalla, Bach, Bahree, Bakhtiari, Behl, et~al.]{abdin2024phi}
Marah Abdin, Sam~Ade Jacobs, Ammar~Ahmad Awan, Jyoti Aneja, Ahmed Awadallah, Hany Awadalla, Nguyen Bach, Amit Bahree, Arash Bakhtiari, Harkirat Behl, et~al.
\newblock Phi-3 technical report: A highly capable language model locally on your phone.
\newblock \emph{arXiv:2404.14219}, 2024.

\bibitem[Achiam et~al.(2023)Achiam, Adler, Agarwal, Ahmad, Akkaya, Aleman, Almeida, Altenschmidt, Altman, Anadkat, et~al.]{achiam2023gpt}
Josh Achiam, Steven Adler, Sandhini Agarwal, Lama Ahmad, Ilge Akkaya, Florencia~Leoni Aleman, Diogo Almeida, Janko Altenschmidt, Sam Altman, Shyamal Anadkat, et~al.
\newblock {GPT-4} technical report.
\newblock \emph{arXiv:2303.08774}, 2023.

\bibitem[Asai et~al.(2023)Asai, Min, Zhong, and Chen]{asai-etal-2023-retrieval}
Akari Asai, Sewon Min, Zexuan Zhong, and Danqi Chen.
\newblock Retrieval-based language models and applications.
\newblock In \emph{ACL}, pages 41--46, 2023.

\bibitem[Biten et~al.(2019)Biten, Tito, Mafla, i~Bigorda, Rusi{\~{n}}ol, Jawahar, Valveny, and Karatzas]{BitenTMBRJVK19}
Ali~Furkan Biten, Rub{\`{e}}n Tito, Andr{\'{e}}s Mafla, Llu{\'{\i}}s~G{\'{o}}mez i Bigorda, Mar{\c{c}}al Rusi{\~{n}}ol, C.~V. Jawahar, Ernest Valveny, and Dimosthenis Karatzas.
\newblock Scene text visual question answering.
\newblock In \emph{ICCV}, pages 4290--4300, 2019.

\bibitem[Borgeaud et~al.(2022)Borgeaud, Mensch, Hoffmann, Cai, Rutherford, Millican, Van Den~Driessche, Lespiau, Damoc, Clark, et~al.]{borgeaud2022improving}
Sebastian Borgeaud, Arthur Mensch, Jordan Hoffmann, Trevor Cai, Eliza Rutherford, Katie Millican, George~Bm Van Den~Driessche, Jean-Baptiste Lespiau, Bogdan Damoc, Aidan Clark, et~al.
\newblock Improving language models by retrieving from trillions of tokens.
\newblock In \emph{ICML}, pages 2206--2240, 2022.

\bibitem[Byeon et~al.(2022)Byeon, Park, Kim, Lee, Baek, and Kim]{kakaobrain2022coyo-700m}
Minwoo Byeon, Beomhee Park, Haecheon Kim, Sungjun Lee, Woonhyuk Baek, and Saehoon Kim.
\newblock Coyo-700m: Image-text pair dataset.
\newblock \url{https://github.com/kakaobrain/coyo-dataset}, 2022.

\bibitem[Chen et~al.(2023)Chen, Pan, Li, Yao, Chao, and Mei]{chen2023retrieval}
Jingwen Chen, Yingwei Pan, Yehao Li, Ting Yao, Hongyang Chao, and Tao Mei.
\newblock Retrieval augmented convolutional encoder-decoder networks for video captioning.
\newblock \emph{TOMCCAP}, pages 1--24, 2023.

\bibitem[Chen et~al.(2022)Chen, Hu, Saharia, and Cohen]{chen2022re}
Wenhu Chen, Hexiang Hu, Chitwan Saharia, and William~W Cohen.
\newblock Re-imagen: Retrieval-augmented text-to-image generator.
\newblock \emph{arXiv:2209.14491}, 2022.

\bibitem[Cho et~al.(2024)Cho, Mahata, Irsoy, He, and Bansal]{cho2024m3docrag}
Jaemin Cho, Debanjan Mahata, Ozan Irsoy, Yujie He, and Mohit Bansal.
\newblock {M3DocRAG}: Multi-modal retrieval is what you need for multi-page multi-document understanding.
\newblock \emph{arXiv:2411.04952}, 2024.

\bibitem[Dai et~al.(2023)Dai, Li, Li, Tiong, Zhao, Wang, Li, Fung, and Hoi]{instructblip}
Wenliang Dai, Junnan Li, Dongxu Li, Anthony Meng~Huat Tiong, Junqi Zhao, Weisheng Wang, Boyang Li, Pascale Fung, and Steven Hoi.
\newblock {InstructBLIP}: Towards general-purpose vision-language models with instruction tuning.
\newblock \emph{arXiv:2305.06500}, 2023.

\bibitem[Dao et~al.(2022)Dao, Fu, Ermon, Rudra, and R{\'e}]{dao2022flashattention}
Tri Dao, Dan Fu, Stefano Ermon, Atri Rudra, and Christopher R{\'e}.
\newblock {FlashAttention}: Fast and memory-efficient exact attention with io-awareness.
\newblock In \emph{NeurIPS}, pages 16344--16359, 2022.

\bibitem[Devlin et~al.(2019)Devlin, Chang, Lee, and Toutanova]{DevlinCLT19}
Jacob Devlin, Ming{-}Wei Chang, Kenton Lee, and Kristina Toutanova.
\newblock {BERT:} pre-training of deep bidirectional transformers for language understanding.
\newblock In \emph{NAACL-HLT}, pages 4171--4186, 2019.

\bibitem[Dong et~al.(2025)Dong, Chang, Goh, Li, Tang, and Liu]{dong2025mmdocir}
Kuicai Dong, Yujing Chang, Xin~Deik Goh, Dexun Li, Ruiming Tang, and Yong Liu.
\newblock {MMDocIR}: Benchmarking multi-modal retrieval for long documents.
\newblock \emph{arXiv:2501.08828}, 2025.

\bibitem[Dong et~al.(2024)Dong, Zhang, Zang, Cao, Wang, Ouyang, Zhang, Duan, Zhang, Li, et~al.]{dong2024internlm}
Xiaoyi Dong, Pan Zhang, Yuhang Zang, Yuhang Cao, Bin Wang, Linke Ouyang, Songyang Zhang, Haodong Duan, Wenwei Zhang, Yining Li, et~al.
\newblock Internlm-xcomposer2-4khd: A pioneering large vision-language model handling resolutions from 336 pixels to 4k hd.
\newblock \emph{arXiv:2404.06512}, 2024.

\bibitem[Douze et~al.(2024)Douze, Guzhva, Deng, Johnson, Szilvasy, Mazaré, Lomeli, Hosseini, and Jégou]{douze2024faiss}
Matthijs Douze, Alexandr Guzhva, Chengqi Deng, Jeff Johnson, Gergely Szilvasy, Pierre-Emmanuel Mazaré, Maria Lomeli, Lucas Hosseini, and Hervé Jégou.
\newblock The faiss library.
\newblock \emph{arXiv:2401.08281}, 2024.

\bibitem[Dubey et~al.(2024)Dubey, Jauhri, Pandey, Kadian, Al-Dahle, Letman, Mathur, Schelten, Yang, Fan, et~al.]{dubey2024llama}
Abhimanyu Dubey, Abhinav Jauhri, Abhinav Pandey, Abhishek Kadian, Ahmad Al-Dahle, Aiesha Letman, Akhil Mathur, Alan Schelten, Amy Yang, Angela Fan, et~al.
\newblock The llama 3 herd of models.
\newblock \emph{arXiv:2407.21783}, 2024.

\bibitem[Faysse et~al.(2024)Faysse, Sibille, Wu, Viaud, Hudelot, and Colombo]{faysse2024colpali}
Manuel Faysse, Hugues Sibille, Tony Wu, Gautier Viaud, C{\'e}line Hudelot, and Pierre Colombo.
\newblock {ColPali}: Efficient document retrieval with vision language models.
\newblock \emph{arXiv:2407.01449}, 2024.

\bibitem[Guu et~al.(2020)Guu, Lee, Tung, Pasupat, and Chang]{guu2020retrieval}
Kelvin Guu, Kenton Lee, Zora Tung, Panupong Pasupat, and Mingwei Chang.
\newblock Retrieval augmented language model pre-training.
\newblock In \emph{ICML}, pages 3929--3938, 2020.

\bibitem[Honnibal and Montani(2017)]{spacy2}
Matthew Honnibal and Ines Montani.
\newblock {spaCy 2}: Natural language understanding with {B}loom embeddings, convolutional neural networks and incremental parsing.
\newblock To appear, 2017.

\bibitem[Hu et~al.(2024)Hu, Xu, Ye, Yan, Zhang, Zhang, Li, Zhang, Jin, Huang, et~al.]{hu2024mplug}
Anwen Hu, Haiyang Xu, Jiabo Ye, Ming Yan, Liang Zhang, Bo Zhang, Chen Li, Ji Zhang, Qin Jin, Fei Huang, et~al.
\newblock mplug-docowl 1.5: Unified structure learning for ocr-free document understanding.
\newblock \emph{arXiv:2403.12895}, 2024.

\bibitem[Hu et~al.(2021)Hu, Shen, Wallis, Allen-Zhu, Li, Wang, Wang, and Chen]{hu2021lora}
Edward~J Hu, Yelong Shen, Phillip Wallis, Zeyuan Allen-Zhu, Yuanzhi Li, Shean Wang, Lu Wang, and Weizhu Chen.
\newblock {LoRA}: Low-rank adaptation of large language models.
\newblock \emph{arXiv:2106.09685}, 2021.

\bibitem[Izacard et~al.(2021)Izacard, Caron, Hosseini, Riedel, Bojanowski, Joulin, and Grave]{izacard2021unsupervised}
Gautier Izacard, Mathilde Caron, Lucas Hosseini, Sebastian Riedel, Piotr Bojanowski, Armand Joulin, and Edouard Grave.
\newblock Unsupervised dense information retrieval with contrastive learning.
\newblock \emph{arXiv:2112.09118}, 2021.

\bibitem[Jiang et~al.(2023)Jiang, Sablayrolles, Mensch, Bamford, Chaplot, de~las Casas, Bressand, Lengyel, Lample, Saulnier, Lavaud, Lachaux, Stock, Scao, Lavril, Wang, Lacroix, and Sayed]{jiang2023mistral7b}
Albert~Q. Jiang, Alexandre Sablayrolles, Arthur Mensch, Chris Bamford, Devendra~Singh Chaplot, Diego de~las Casas, Florian Bressand, Gianna Lengyel, Guillaume Lample, Lucile Saulnier, Lélio~Renard Lavaud, Marie-Anne Lachaux, Pierre Stock, Teven~Le Scao, Thibaut Lavril, Thomas Wang, Timothée Lacroix, and William~El Sayed.
\newblock Mistral 7b.
\newblock \emph{arXiv:2310.06825}, 2023.

\bibitem[Jiang et~al.(2024)Jiang, Sablayrolles, Roux, Mensch, Savary, Bamford, Chaplot, Casas, Hanna, Bressand, et~al.]{jiang2024mixtral}
Albert~Q Jiang, Alexandre Sablayrolles, Antoine Roux, Arthur Mensch, Blanche Savary, Chris Bamford, Devendra~Singh Chaplot, Diego de~las Casas, Emma~Bou Hanna, Florian Bressand, et~al.
\newblock Mixtral of experts.
\newblock \emph{arXiv:2401.04088}, 2024.

\bibitem[Kamalloo et~al.(2023)Kamalloo, Thakur, Lassance, Ma, Yang, and Lin]{kamalloo2023resources}
Ehsan Kamalloo, Nandan Thakur, Carlos Lassance, Xueguang Ma, Jheng-Hong Yang, and Jimmy Lin.
\newblock Resources for brewing beir: Reproducible reference models and an official leaderboard, 2023.

\bibitem[Koizumi et~al.(2020)Koizumi, Ohishi, Niizumi, Takeuchi, and Yasuda]{koizumi2020audio}
Yuma Koizumi, Yasunori Ohishi, Daisuke Niizumi, Daiki Takeuchi, and Masahiro Yasuda.
\newblock Audio captioning using pre-trained large-scale language model guided by audio-based similar caption retrieval.
\newblock \emph{arXiv:2012.07331}, 2020.

\bibitem[Kweon et~al.(2023)Kweon, Kwon, Cho, Jo, and Choi]{kweon-etal-2023-open}
Sunjun Kweon, Yeonsu Kwon, Seonhee Cho, Yohan Jo, and Edward Choi.
\newblock {Open-WikiTable} : Dataset for open domain question answering with complex reasoning over table.
\newblock In \emph{Findings of ACL}, pages 8285--8297, 2023.

\bibitem[Landeghem et~al.(2023)Landeghem, Tito, Borchmann, Pietruszka, J{\'o}ziak, Powalski, Jurkiewicz, Coustaty, Ackaert, Valveny, et~al.]{landeghem2023document}
Jordy Landeghem, Rub{\'e}n Tito, {\L}ukasz Borchmann, Micha{\l} Pietruszka, Pawe{\l} J{\'o}ziak, Rafa{\l} Powalski, Dawid Jurkiewicz, Micka{\"e}l Coustaty, Bertrand Ackaert, Ernest Valveny, et~al.
\newblock Document understanding dataset and evaluation (dude).
\newblock In \emph{ICCV}, pages 19528--19540, 2023.

\bibitem[Lauren{\c{c}}on et~al.(2024)Lauren{\c{c}}on, Marafioti, Sanh, and Tronchon]{laurenccon2024building}
Hugo Lauren{\c{c}}on, Andr{\'e}s Marafioti, Victor Sanh, and L{\'e}o Tronchon.
\newblock Building and better understanding vision-language models: insights and future directions.
\newblock \emph{arXiv:2408.12637}, 2024.

\bibitem[Lee et~al.(2024)Lee, Roy, Xu, Raiman, Shoeybi, Catanzaro, and Ping]{lee2024nv}
Chankyu Lee, Rajarshi Roy, Mengyao Xu, Jonathan Raiman, Mohammad Shoeybi, Bryan Catanzaro, and Wei Ping.
\newblock {Nv-Embed}: Improved techniques for training llms as generalist embedding models.
\newblock \emph{arXiv:2405.17428}, 2024.

\bibitem[Lewis et~al.(2020)Lewis, Perez, Piktus, Petroni, Karpukhin, Goyal, K{\"u}ttler, Lewis, Yih, Rockt{\"a}schel, et~al.]{lewis2020retrieval}
Patrick Lewis, Ethan Perez, Aleksandra Piktus, Fabio Petroni, Vladimir Karpukhin, Naman Goyal, Heinrich K{\"u}ttler, Mike Lewis, Wen-tau Yih, Tim Rockt{\"a}schel, et~al.
\newblock Retrieval-augmented generation for knowledge-intensive nlp tasks.
\newblock In \emph{NIPS}, pages 9459--9474, 2020.

\bibitem[Li et~al.(2022)Li, Li, Xiong, and Hoi]{li2022blip}
Junnan Li, Dongxu Li, Caiming Xiong, and Steven Hoi.
\newblock Blip: Bootstrapping language-image pre-training for unified vision-language understanding and generation.
\newblock In \emph{ICML}, pages 12888--12900, 2022.

\bibitem[Li et~al.(2023{\natexlab{a}})Li, Li, Savarese, and Hoi]{li2023blip2}
Junnan Li, Dongxu Li, Silvio Savarese, and Steven Hoi.
\newblock {BLIP-2:} bootstrapping language-image pre-training with frozen image encoders and large language models.
\newblock In \emph{ICML}, pages 19730--19742, 2023{\natexlab{a}}.

\bibitem[Li et~al.(2023{\natexlab{b}})Li, Zhang, Zhang, Long, Xie, and Zhang]{li2023towards}
Zehan Li, Xin Zhang, Yanzhao Zhang, Dingkun Long, Pengjun Xie, and Meishan Zhang.
\newblock Towards general text embeddings with multi-stage contrastive learning.
\newblock \emph{arXiv:2308.03281}, 2023{\natexlab{b}}.

\bibitem[Liu et~al.(2023)Liu, Li, Wu, and Lee]{liu2023llava}
Haotian Liu, Chunyuan Li, Qingyang Wu, and Yong~Jae Lee.
\newblock Visual instruction tuning.
\newblock \emph{arXiv:2304.08485}, 2023.

\bibitem[Loshchilov and Hutter(2017)]{loshchilov2017decoupled}
Ilya Loshchilov and Frank Hutter.
\newblock Decoupled weight decay regularization.
\newblock \emph{arXiv:1711.05101}, 2017.

\bibitem[Ma et~al.(2024{\natexlab{a}})Ma, Lin, Li, Chen, and Lin]{ma2024unifying}
Xueguang Ma, Sheng-Chieh Lin, Minghan Li, Wenhu Chen, and Jimmy Lin.
\newblock Unifying multimodal retrieval via document screenshot embedding.
\newblock \emph{arXiv:2406.11251}, 2024{\natexlab{a}}.

\bibitem[Ma et~al.(2024{\natexlab{b}})Ma, Zhuang, Koopman, Zuccon, Chen, and Lin]{ma2024visa}
Xueguang Ma, Shengyao Zhuang, Bevan Koopman, Guido Zuccon, Wenhu Chen, and Jimmy Lin.
\newblock {VISA}: Retrieval augmented generation with visual source attribution.
\newblock \emph{arXiv:2412.14457}, 2024{\natexlab{b}}.

\bibitem[Maekawa et~al.(2024)Maekawa, Iso, Gurajada, and Bhutani]{maekawa-etal-2024-retrieval}
Seiji Maekawa, Hayate Iso, Sairam Gurajada, and Nikita Bhutani.
\newblock Retrieval helps or hurts? a deeper dive into the efficacy of retrieval augmentation to language models.
\newblock In \emph{NAACL}, pages 5506--5521, 2024.

\bibitem[Mallen et~al.(2023)Mallen, Asai, Zhong, Das, Khashabi, and Hajishirzi]{mallen-etal-2023-trust}
Alex Mallen, Akari Asai, Victor Zhong, Rajarshi Das, Daniel Khashabi, and Hannaneh Hajishirzi.
\newblock When not to trust language models: Investigating effectiveness of parametric and non-parametric memories.
\newblock In \emph{ACL}, pages 9802--9822, 2023.

\bibitem[Masry et~al.(2022)Masry, Do, Tan, Joty, and Hoque]{masry-etal-2022-chartqa}
Ahmed Masry, Xuan~Long Do, Jia~Qing Tan, Shafiq Joty, and Enamul Hoque.
\newblock {C}hart{QA}: A benchmark for question answering about charts with visual and logical reasoning.
\newblock In \emph{Findings of ACL}, pages 2263--2279, 2022.

\bibitem[Mathew et~al.(2021)Mathew, Karatzas, and Jawahar]{Mathew_2021_WACV}
Minesh Mathew, Dimosthenis Karatzas, and C.~V. Jawahar.
\newblock {DocVQA}: A dataset for vqa on document images.
\newblock In \emph{WACV}, pages 2200--2209, 2021.

\bibitem[Mathew et~al.(2022)Mathew, Bagal, Tito, Karatzas, Valveny, and Jawahar]{Mathew_2022_WACV}
Minesh Mathew, Viraj Bagal, Rub\`en Tito, Dimosthenis Karatzas, Ernest Valveny, and C.V. Jawahar.
\newblock {InfographicVQA}.
\newblock In \emph{WACV}, pages 1697--1706, 2022.

\bibitem[Oord et~al.(2018)Oord, Li, and Vinyals]{oord2018representation}
Aaron van~den Oord, Yazhe Li, and Oriol Vinyals.
\newblock Representation learning with contrastive predictive coding.
\newblock \emph{arXiv:1807.03748}, 2018.

\bibitem[Parvez et~al.(2021)Parvez, Ahmad, Chakraborty, Ray, and Chang]{parvez2021retrieval}
Md~Rizwan Parvez, Wasi~Uddin Ahmad, Saikat Chakraborty, Baishakhi Ray, and Kai-Wei Chang.
\newblock Retrieval augmented code generation and summarization.
\newblock \emph{arXiv:2108.11601}, 2021.

\bibitem[Qi et~al.(2022)Qi, Lv, Li, Liu, Zhang, She, Wu, Wang, and Liu]{qi-etal-2022-dureadervis}
Le Qi, Shangwen Lv, Hongyu Li, Jing Liu, Yu Zhang, Qiaoqiao She, Hua Wu, Haifeng Wang, and Ting Liu.
\newblock $\textrm{DuReader}_{\textrm{vis}}$: A {C}hinese dataset for open-domain document visual question answering.
\newblock In \emph{Findings of ACL}, pages 1338--1351, 2022.

\bibitem[Radford et~al.(2021)Radford, Kim, Hallacy, Ramesh, Goh, Agarwal, Sastry, Askell, Mishkin, Clark, et~al.]{radford2021learning}
Alec Radford, Jong~Wook Kim, Chris Hallacy, Aditya Ramesh, Gabriel Goh, Sandhini Agarwal, Girish Sastry, Amanda Askell, Pamela Mishkin, Jack Clark, et~al.
\newblock Learning transferable visual models from natural language supervision.
\newblock In \emph{ICML}, pages 8748--8763, 2021.

\bibitem[Raffel et~al.(2020)Raffel, Shazeer, Roberts, Lee, Narang, Matena, Zhou, Li, and Liu]{RaffelSRLNMZLL20}
Colin Raffel, Noam Shazeer, Adam Roberts, Katherine Lee, Sharan Narang, Michael Matena, Yanqi Zhou, Wei Li, and Peter~J. Liu.
\newblock Exploring the limits of transfer learning with a unified text-to-text transformer.
\newblock \emph{JMLR}, 21\penalty0 (140):\penalty0 1--67, 2020.

\bibitem[Ram et~al.(2023)Ram, Levine, Dalmedigos, Muhlgay, Shashua, Leyton-Brown, and Shoham]{ram2023context}
Ori Ram, Yoav Levine, Itay Dalmedigos, Dor Muhlgay, Amnon Shashua, Kevin Leyton-Brown, and Yoav Shoham.
\newblock In-context retrieval-augmented language models.
\newblock \emph{TACL}, pages 1316--1331, 2023.

\bibitem[Ramos et~al.(2023{\natexlab{a}})Ramos, Elliott, and Martins]{ramos-etal-2023-retrieval}
Rita Ramos, Desmond Elliott, and Bruno Martins.
\newblock Retrieval-augmented image captioning.
\newblock In \emph{EACL}, pages 3666--3681, 2023{\natexlab{a}}.

\bibitem[Ramos et~al.(2023{\natexlab{b}})Ramos, Martins, Elliott, and Kementchedjhieva]{ramos2023smallcap}
Rita Ramos, Bruno Martins, Desmond Elliott, and Yova Kementchedjhieva.
\newblock Smallcap: lightweight image captioning prompted with retrieval augmentation.
\newblock In \emph{CVPR}, pages 2840--2849, 2023{\natexlab{b}}.

\bibitem[Robertson et~al.(2009)Robertson, Zaragoza, et~al.]{robertson2009probabilistic}
Stephen Robertson, Hugo Zaragoza, et~al.
\newblock The probabilistic relevance framework: Bm25 and beyond.
\newblock \emph{Foundations and Trends{\textregistered} in Information Retrieval}, 3\penalty0 (4):\penalty0 333--389, 2009.

\bibitem[Seo et~al.(2024)Seo, Hong, Jang, Kim, Kwak, Lee, and Kim]{seo2024retrieval}
Junyoung Seo, Susung Hong, Wooseok Jang, In{\`e}s~Hyeonsu Kim, Minseop Kwak, Doyup Lee, and Seungryong Kim.
\newblock Retrieval-augmented score distillation for text-to-3d generation.
\newblock \emph{arXiv:2402.02972}, 2024.

\bibitem[Smith(2007)]{smith2007overview}
Ray Smith.
\newblock An overview of the tesseract ocr engine.
\newblock In \emph{ICDAR}, pages 629--633, 2007.

\bibitem[Suzgun et~al.(2022)Suzgun, Scales, Sch{\"a}rli, Gehrmann, Tay, Chung, Chowdhery, Le, Chi, Zhou, et~al.]{suzgun2022challenging}
Mirac Suzgun, Nathan Scales, Nathanael Sch{\"a}rli, Sebastian Gehrmann, Yi Tay, Hyung~Won Chung, Aakanksha Chowdhery, Quoc~V Le, Ed~H Chi, Denny Zhou, et~al.
\newblock Challenging big-bench tasks and whether chain-of-thought can solve them.
\newblock \emph{arXiv:2210.09261}, 2022.

\bibitem[Tanaka et~al.(2021)Tanaka, Nishida, and Yoshida]{DBLP:conf/aaai/TanakaNY21}
Ryota Tanaka, Kyosuke Nishida, and Sen Yoshida.
\newblock {VisualMRC}: Machine reading comprehension on document images.
\newblock In \emph{AAAI}, pages 13878--13888, 2021.

\bibitem[Tanaka et~al.(2023)Tanaka, Nishida, Nishida, Hasegawa, Saito, and Saito]{SlideVQA2023}
Ryota Tanaka, Kyosuke Nishida, Kosuke Nishida, Taku Hasegawa, Itsumi Saito, and Kuniko Saito.
\newblock {SlideVQA}: A dataset for document visual question answering on multiple images.
\newblock In \emph{AAAI}, pages 13636--13645, 2023.

\bibitem[Tanaka et~al.(2024)Tanaka, Iki, Nishida, Saito, and Suzuki]{tanaka2024instructdoc}
Ryota Tanaka, Taichi Iki, Kyosuke Nishida, Kuniko Saito, and Jun Suzuki.
\newblock Instructdoc: A dataset for zero-shot generalization of visual document understanding with instructions.
\newblock In \emph{AAAI}, pages 19071--19079, 2024.

\bibitem[Wang et~al.(2022)Wang, Yang, Huang, Jiao, Yang, Jiang, Majumder, and Wei]{wang2022text}
Liang Wang, Nan Yang, Xiaolong Huang, Binxing Jiao, Linjun Yang, Daxin Jiang, Rangan Majumder, and Furu Wei.
\newblock Text embeddings by weakly-supervised contrastive pre-training.
\newblock \emph{arXiv:2212.03533}, 2022.

\bibitem[Wang et~al.(2024)Wang, Yang, Huang, Yang, Majumder, and Wei]{wang-etal-2024-improving-text}
Liang Wang, Nan Yang, Xiaolong Huang, Linjun Yang, Rangan Majumder, and Furu Wei.
\newblock Improving text embeddings with large language models.
\newblock In \emph{ACL}, pages 11897--11916, 2024.

\bibitem[Xu et~al.(2024)Xu, Huang, Hou, Chen, Zhang, Feng, and Xie]{xu2024retrieval}
Jilan Xu, Yifei Huang, Junlin Hou, Guo Chen, Yuejie Zhang, Rui Feng, and Weidi Xie.
\newblock Retrieval-augmented egocentric video captioning.
\newblock In \emph{CVPR}, pages 13525--13536, 2024.

\bibitem[Yang et~al.(2024)Yang, Liu, Huang, Weng, and Meng]{yang2024instructtts}
Dongchao Yang, Songxiang Liu, Rongjie Huang, Chao Weng, and Helen Meng.
\newblock Instructtts: Modelling expressive tts in discrete latent space with natural language style prompt.
\newblock \emph{TASLP}, pages 2913--2925, 2024.

\bibitem[Yao et~al.(2024)Yao, Yu, Zhang, Wang, Cui, Zhu, Cai, Li, Zhao, He, Chen, Zhou, Zou, Zhang, Hu, Zheng, Zhou, Cai, Han, Zeng, Li, Liu, and Sun]{yao2024minicpmvgpt4vlevelmllm}
Yuan Yao, Tianyu Yu, Ao Zhang, Chongyi Wang, Junbo Cui, Hongji Zhu, Tianchi Cai, Haoyu Li, Weilin Zhao, Zhihui He, Qianyu Chen, Huarong Zhou, Zhensheng Zou, Haoye Zhang, Shengding Hu, Zhi Zheng, Jie Zhou, Jie Cai, Xu Han, Guoyang Zeng, Dahai Li, Zhiyuan Liu, and Maosong Sun.
\newblock Minicpm-v: A gpt-4v level mllm on your phone.
\newblock \emph{arXiv:2408.01800}, 2024.

\bibitem[Yasunaga et~al.(2023)Yasunaga, Aghajanyan, Shi, James, Leskovec, Liang, Lewis, Zettlemoyer, and Yih]{yasunaga2023retrieval}
Michihiro Yasunaga, Armen Aghajanyan, Weijia Shi, Rich James, Jure Leskovec, Percy Liang, Mike Lewis, Luke Zettlemoyer, and Wen-tau Yih.
\newblock Retrieval-augmented multimodal language modeling.
\newblock In \emph{ICML}, pages 39755--39769, 2023.

\bibitem[Ye et~al.(2023)Ye, Hu, Xu, Ye, Yan, Xu, Li, Tian, Qian, Zhang, Jin, He, Lin, and Huang]{ye-etal-2023-ureader}
Jiabo Ye, Anwen Hu, Haiyang Xu, Qinghao Ye, Ming Yan, Guohai Xu, Chenliang Li, Junfeng Tian, Qi Qian, Ji Zhang, Qin Jin, Liang He, Xin Lin, and Fei Huang.
\newblock {UR}eader: Universal {OCR}-free visually-situated language understanding with multimodal large language model.
\newblock In \emph{EMNLP Findings}, pages 2841--2858, 2023.

\bibitem[Yu et~al.(2024)Yu, Tang, Xu, Cui, Ran, Yan, Liu, Wang, Han, Liu, et~al.]{yu2024visrag}
Shi Yu, Chaoyue Tang, Bokai Xu, Junbo Cui, Junhao Ran, Yukun Yan, Zhenghao Liu, Shuo Wang, Xu Han, Zhiyuan Liu, et~al.
\newblock {VisRAG}: Vision-based retrieval-augmented generation on multi-modality documents.
\newblock \emph{arXiv:2410.10594}, 2024.

\bibitem[Zhai et~al.(2023)Zhai, Mustafa, Kolesnikov, and Beyer]{zhai2023sigmoid}
Xiaohua Zhai, Basil Mustafa, Alexander Kolesnikov, and Lucas Beyer.
\newblock Sigmoid loss for language image pre-training.
\newblock In \emph{ICCV}, pages 11975--11986, 2023.

\bibitem[Zhang et~al.(2023{\natexlab{a}})Zhang, Hu, Zhang, Hu, and Jin]{zhang2023mpmqa}
Liang Zhang, Anwen Hu, Jing Zhang, Shuo Hu, and Qin Jin.
\newblock {MPMQA}: multimodal question answering on product manuals.
\newblock In \emph{AAAI}, pages 13958--13966, 2023{\natexlab{a}}.

\bibitem[Zhang et~al.(2023{\natexlab{b}})Zhang, Guo, Pan, Cai, Hong, Li, Yang, and Liu]{zhang2023remodiffuse}
Mingyuan Zhang, Xinying Guo, Liang Pan, Zhongang Cai, Fangzhou Hong, Huirong Li, Lei Yang, and Ziwei Liu.
\newblock Remodiffuse: Retrieval-augmented motion diffusion model.
\newblock In \emph{ICCV}, pages 364--373, 2023{\natexlab{b}}.

\bibitem[Zhou et~al.(2022)Zhou, Alon, Xu, Wang, Jiang, and Neubig]{zhou2022docprompting}
Shuyan Zhou, Uri Alon, Frank~F Xu, Zhiruo Wang, Zhengbao Jiang, and Graham Neubig.
\newblock Docprompting: Generating code by retrieving the docs.
\newblock \emph{arXiv:2207.05987}, 2022.

\bibitem[Zhu et~al.(2022)Zhu, Lei, Feng, Wang, Zhang, and Chua]{zhu2022towards}
Fengbin Zhu, Wenqiang Lei, Fuli Feng, Chao Wang, Haozhou Zhang, and Tat-Seng Chua.
\newblock Towards complex document understanding by discrete reasoning.
\newblock In \emph{ACMM}, pages 4857--4866, 2022.

\end{thebibliography}
